\documentclass{article}

\usepackage{arxiv}
\usepackage[utf8]{inputenc} 
\usepackage[T1]{fontenc}    
\usepackage{hyperref}       
\usepackage{url}            
\usepackage{booktabs}       
\usepackage{amsfonts}       
\usepackage{nicefrac}       
\usepackage{microtype}      
\usepackage{lipsum}
\usepackage{graphicx}       

\usepackage{fancyhdr}       
\usepackage{anyfontsize}    

\usepackage[onehalfspacing]{setspace}   
\usepackage{natbib}
\usepackage{amssymb}
\usepackage{xcolor}                     
\usepackage[english]{babel}
\usepackage{mathtools}                  
\usepackage{multirow}                   
\usepackage{makecell}                   
\usepackage{cleveref}                   
\usepackage[acronym, symbols]{glossaries}

\usepackage{comment}                    

\newglossarystyle{tree-cap-desc}{%
  \setglossarystyle{tree}
}

\makeatother

\makeglossaries

\setglossarystyle{tree}      

\renewcommand*{\glsnamefont}[1]{#1\hspace{1em}} 

\newacronym{SOTA}{SOTA}{state of the art}
\newacronym{IOU}{IoU}{intersection over union}

\newacronym{CRS}{CRS}{coordinate reference system}
\newacronym{NED}{NED}{North-East-Down}
\newacronym{ECEF}{ECEF}{Earth-centered Earth-fixed}
\newacronym{SNAME}{SNAME}{Society of Naval Architects and Marine Engineers}

\newacronym{GNSS}{GNSS}{global navigation satellite system}
\newacronym{VAN}{VAN}{vision-aided navigation}
\newacronym{SLAM}{SLAM}{simultaneous localization and mapping}

\newacronym{SFM}{SFM}{structure-from-motion}
\newacronym{MVS}{MVS}{multi-view stereo}
\newacronym{SFM-MVS}{SFM-MVS}{structure-from-motion multi-view stereo}

\newacronym{AUV}{AUV}{autonomous underwater vehicle}
\newacronym{ROV}{ROV}{remotely operated vehicle}
\newacronym{DVL}{DVL}{Doppler velocity log}
\newacronym{APS}{APS}{acoustic positioning system}
\newacronym{USBL}{USBL}{ultra-short baseline}

\newacronym{VPR}{VPR}{visual place recognition}
\newacronym{CNN}{CNN}{convolutional neural network}
\newacronym{VIT}{ViT}{vision transformer}
\newacronym{VLAD}{VLAD}{vector of locally aggregated descriptors}
\newacronym{IR}{IR}{information retrieval}

\newacronym{ACFR}{ACFR}{Australian Centre of Field Robotics}
\newacronym{IMOS}{IMOS}{Integrated Marine Observing System}

\newacronym{FPFH}{FPFH}{fast point feature histogram}
\newacronym{ICP}{ICP}{iterative closest point}
\newacronym{RANSAC}{RANSAC}{random sample consensus}

\newacronym{ALQ}{ALQ}{average link per query}

\newcommand{\norm}[1]{\left\lVert#1\right\rVert}
\newcommand{\transpose}{\top}
\newcommand{\homogeneous}[1]{\tilde{#1}}

\newcommand{\indexed}[2]{\prescript{#2}{}{#1}} 

\newcommand{\transformRigid}[2]{\prescript{#2}{#1}{\mathbf{T}}}

\newcommand{\cameraOriginVec}{\mathbf{o}}
\newcommand{\principalPointX}{u_{0}}
\newcommand{\principalPointY}{v_{0}}
\newcommand{\focalLengthX}{\alpha_{x}}
\newcommand{\focalLengthY}{\alpha_{y}}
\newcommand{\fieldOfView}{\mathrm{FOV}}

\newcommand{\intrinsicMatrix}{\mathbf{K}} 
\newcommand{\range}{z}

\newcommand{\objectCoordVec}{\mathbf{p}}
\newcommand{\objectCoordVecHomo}{\homogeneous{\objectCoordVec}}

\newcommand{\pixelCoordX}{u}
\newcommand{\pixelCoordY}{v}
\newcommand{\pixelCoordVec}{\mathbf{u}}
\newcommand{\pixelCoordVecHomo}{\homogeneous{\pixelCoordVec}}

\newcommand{\footprintSet}{\mathcal{P}}
\newcommand{\footprintSetProj}{\bar{\footprintSet}}
\newcommand{\footprintIou}{\mathrm{IoU}}

\newcommand{\imageColorChannel}{\lambda}
\newcommand{\imageDesiredMean}{\mu_{y}}
\newcommand{\imageDesiredStd}{\sigma_{y}}

\newcommand{\lieGroupSE}[1]{\mathrm{SE}\left(#1\right)}
\newcommand{\lieGroupSim}[1]{\mathrm{Sim}\left(#1\right)}

\newcommand{\spatialUnion}{\cup_{s}}
\newcommand{\spatialIntersect}{\cap_{s}}

\DeclareMathOperator{\SpatialIntersectFun}{SpatialIntersect}
\DeclareMathOperator{\SpatialUnionFun}{SpatialUnion}
\DeclareMathOperator{\SpatialAreaFun}{Area}

\DeclareMathOperator{\topKFun}{\mathrm{TopK}}

\graphicspath{ {./images/} }

\title{
    Long-Term Visual Localization in Dynamic Benthic Environments: A Dataset, Footprint-Based Ground Truth, and Visual Place Recognition Benchmark
}

\author{
    Martin Kvisvik Larsen\\
    Department of Marine Technology\\
    Norwegian University of Science and Technology\\
    Trondheim, Norway\\
    \texttt{martin.k.larsen@ntnu.no}\\
    \And
    Oscar Pizarro \\
    Department of Marine Technology\\
    Norwegian University of Science and Technology\\
    Trondheim, Norway\\
    \texttt{oscar.pizarro@ntnu.no}\\
}

\begin{document}

\maketitle

\begin{abstract}
Long-term visual localization has the potential to reduce cost and improve mapping quality in optical benthic monitoring with autonomous underwater vehicles (AUVs). Despite this potential, long-term visual localization in benthic environments remains understudied, primarily due to the lack of curated datasets for benchmarking. Moreover, limited georeferencing accuracy and image footprints necessitate precise geometric information for accurate ground-truthing. In this work, we address these gaps by presenting a curated dataset for long-term visual localization in benthic environments and a novel method to ground-truth visual localization results for near-nadir underwater imagery. Our dataset comprises georeferenced AUV imagery from five benthic reference sites, revisited over periods up to six years, and includes raw and color-corrected stereo imagery, camera calibrations, and sub-decimeter registered camera poses. To our knowledge, this is the first curated underwater dataset for long-term visual localization spanning multiple sites and photic-zone habitats. Our ground-truthing method estimates 3D seafloor image footprints and links camera views with overlapping footprints, ensuring that ground-truth links reflect shared visual content. Building on this dataset and ground truth, we benchmark eight state-of-the-art visual place recognition (VPR) methods and find that Recall@K is significantly lower on our dataset than on established terrestrial and underwater benchmarks. Finally, we compare our footprint-based ground truth to a traditional location-based ground truth and show that distance-threshold ground-truthing can overestimate VPR Recall@K at sites with rugged terrain and altitude variations. Together, the curated dataset, ground-truthing method, and VPR benchmark provide a stepping stone for advancing long-term visual localization in dynamic benthic environments.
\end{abstract}

\keywords{
    Benthic environments 
    \and Dataset 
    \and Ground truth
    \and Long-Term visual localization
    \and Visual place recognition,
    \and VPR
}


\section{Introduction}
\label{sec:introduction}

Optical monitoring of benthic habitats using robotic platforms is changing marine ecological research by providing high-resolution, repeatable imagery over broad areas and depths where human access is limited~\citep{misiuk_benthic_2024}. Cameras mounted on \glspl{AUV} or \glspl{ROV} enable detailed, non-destructive mapping of species and habitat structures, supporting quantitative and systematic seafloor assessments that were previously unattainable. Due to the unavailability of \gls{GNSS}, underwater robotic platforms often depend on \glspl{APS} for navigation and image georeferencing~\citep{wu_survey_2019}. However, deploying \glspl{APS} involves substantial costs and logistical overhead, requiring auxiliary infrastructure such as topside and subsea transducers, a support vessel for tracking, and specialized personnel for setup and calibration. In addition, \glspl{APS} are highly sensitive to calibration and alignment errors~\citep{chen_-situ_2008, jinwu_study_2018, zhu_calibration_2020}. For benthic repeat surveys conducted months or years apart, calibrations can be invalidated due to sensor drift or misalignment after hardware reinstallation. Consequently, image georeferencing accuracy across visits to the same benthic location is typically limited to several meters~\citep{bryson_automated_2013, larsen_geometric_2023}.

Long-term visual localization provides a means for underwater robots to estimate their position by recognizing distinctive features of the seafloor observed during previous surveys. In this approach, the visual structure of the benthic environment itself serves as a natural reference for \gls{VAN}~\citep{mahon_efficient_2008, eustice_visually_2008, zhang_autonomous_2023}, reducing the need for continuous external tracking. By leveraging visual cues rather than depending exclusively on \glspl{APS}, underwater platforms can operate with reduced reliance on costly and logistically demanding infrastructure. This capability has the potential to enable benthic repeat surveys using simpler sensor platforms and less extensive support equipment, thereby improving operational efficiency and lowering overall survey costs. 

In addition to logistical and cost advantages, long-term visual localization also enhances the spatial quality and consistency of benthic imaging datasets by enabling robots to precisely relocalize within previously mapped areas. This capability allows underwater robots to align new surveys with historical ones in real time, ensuring consistent coverage and maintaining sufficient image overlap across revisits. Furthermore, precise relocalization supports centimeter-scale registration of imagery between surveys, facilitating the detection and quantification of small-scale habitat changes such as species turnover or morphological alterations. For instance, temporally registered imagery has been shown to enable attribution of structural habitat changes to specific taxonomic groups, underscoring the ecological value of spatially consistent optical monitoring datasets~\citep{ferrari_quantifying_2016}. Consequently, visual localization helps improve mapping quality and registration across revisits, thereby enhancing the ability of optical benthic monitoring programs to detect ecological changes and to inform evidence-based habitat management.

Despite its promise, long-term visual localization in benthic environments remains challenging. Images captured by underwater robots are degraded by strong light attenuation, backscatter, and non-uniform illumination, all of which diminish image quality and limit the effective sensing range and image footprint~\citep{song_optical_2022}. Additionally, the temporal dynamics of benthic habitats, including among other the growth, decay, and death of sessile organisms, sediment disturbances, and collapse of structural features, further challenge the robustness of established visual localization methods when applied to imagery collected over months or years~\citep{bryson_automated_2013}. Advances in long-term visual localization research for terrestrial environments have largely been driven by the availability of curated datasets that support systematic method development and evaluation~\citep{toft_long-term_2022}. In contrast, similar datasets from underwater environments remain rare. To date, the only curated underwater dataset featuring multiple revisits over extended periods is the Eiffel Tower dataset~\citep{boittiaux_eiffel_2023}, which consists of \gls{ROV} imagery of a hydrothermal vent in the Lucky Strike vent field on the Mid-Atlantic Ridge at \textasciitilde 1700 meters depth. The dataset consists of imagery collected during visits in 2015, 2016, 2018, and 2020, and covers an area of \textasciitilde 40$\times$40 meters at a spatial resolution of \textasciitilde 2.4 mm / pixel.

In this work, we present a curated imaging \gls{AUV} dataset designed for long-term visual localization research in benthic environments. The dataset addresses the lack of resources capturing the natural temporal variability of photic-zone habitats, providing a basis for developing robust methodologies for underwater visual localization. The dataset consists of repeat surveys at five benthic reference sites located at depths between \textasciitilde18 and \textasciitilde45 meters. Each site spans approximately 35$\times$35 meters and is mapped at a spatial resolution of \textasciitilde1.3 mm/pixel. The reference sites cover diverse seafloor types, including sparse and dense coral reefs, soft sediment bottoms, rock reefs, and boulder reefs. For each site, we provide raw and color-corrected imagery, along with geometrically registered camera poses. Color corrections are computed using a gray-world approach, assuming gray mean intensity across multiple images. Unlike the Eiffel Tower dataset~\citep{boittiaux_eiffel_2023}, which captures a hydrothermal vent at \textasciitilde1600 meters depth, our dataset covers photic-zone environments that experience strong temporal dynamics driven by events such as extreme storms and marine heatwaves~\citep{wong_systematic_2023, perkins_temporal_2025}. In contrast, hydrothermal vent communities at the Mid-Atlantic Ridge are considered stable over decadal timescales in the absence of major eruption events~\citep{glover_temporal_2010}. These contrasting conditions make photic-zone habitats a more challenging testbed for long-term visual localization. Building on this dataset, we present a benchmark of eight \gls{SOTA} \gls{VPR} models to evaluate their performance in dynamic benthic environments. Evaluation of visual localization methods for near-nadir (i.e., ``down-looking'') \gls{AUV} imagery brings unique challenges, as traditional ground-truthing based on location proximity thresholds may be inadequate where the camera's sensing range is similar to the relief of the seafloor. Previous studies~\citep{larsen_geometric_2023, gorry_image-based_2025} typically classified true localizations by requiring the location of the query and database images to be within a fixed distance. However, empirical evidence indicates that this approach can lead to incorrect assessments when the sensing range changes throughout the dataset, either due to sudden changes in seafloor relief or variations in vehicle altitude~\citep{larsen_geometric_2023}. To address these limitations, we propose an evaluation method that leverages range data to estimate the 3D coordinate of the image corners onto the seafloor, allowing a geometric computation of each image's footprint. True localizations are then classified by footprint overlap between the query and database images, ensuring that positive matches are classified based on common visual content, rather than location proximity alone.

Collectively, our curated dataset, ground-truthing approach, and comprehensive \gls{VPR} benchmark are intended to support and accelerate the development of models and methods for long-term visual localization in dynamic benthic environments. Our main contributions are:

\begin{itemize}
    \item A curated dataset of \gls{AUV} imagery for long-term visual localization in benthic environments, providing geometrically registered camera poses, camera calibrations, raw and color-corrected images. The dataset is the first curated dataset for this purpose that covers habitats from the photic zone and includes five benthic reference sites with diverse seafloor types revisited over periods up to 6 years.
    \item A ground-truthing approach for visual localization results for near-nadir \gls{AUV} imagery, using overlapping image footprints to classify correct localization and eliminating the need for location proximity thresholds.
    \item A comprehensive benchmark of established \gls{VPR} methods assessed on the new dataset, advancing the development and evaluation of robust visual localization techniques for repeatable optical mapping of benthic habitats.
    \item We show that \gls{VPR} performance metrics are affected by the choice of ground truth definition by comparing our footprint-based ground truth to a location-based ground truth. We show that our footprint-based ground truth yields consistent results across sites, and that the location-based ground truth overestimates model performance for sites with rugged terrain or visits with large altitude variations.
\end{itemize}


\section{Methods}
\label{sec:methods}

We describe data selection, image color correction, geometric processing, footprint estimation, and the \gls{VPR} evaluation protocol.


\subsection{Data Source and Dataset Overview}
\label{sec:method_data_source}

Long-term, georeferenced optical monitoring datasets of the seafloor are rare, particularly those that repeatedly image fixed reference sites at high spatial resolution over multiple years. Australia’s~\gls{IMOS} addresses this gap through a national benthic imaging program that acquires optical~\glspl{AUV} dataset along the continental shelf, targeting reef habitats and other structurally complex seafloor types that are sensitive to environmental change~\citep{williams_monitoring_2012}. Within this program, the~\gls{IMOS}~\gls{AUV} Facility operates a network of benthic reference sites that are revisited on multi-year timescales, following a nested sampling design that combines broad, sparse grids with dense “mow-the-lawn” grids on the order of 25$\times$25~m to support repeat mapping and quantitative change detection~\citep{pizarro_benthic_2013}. The imagery in our dataset was collected by AUV Sirius, the primary benthic imaging platform of the~\gls{IMOS} AUV Facility~\citep{williams_monitoring_2012}. AUV Sirius is a mid-size SeaBED-class vehicle designed for low-speed, near-bottom surveys, with passive stability in pitch and roll to enable consistent, near-nadir optical imaging~\citep{pizarro_benthic_2013}. The vehicle executes pre-planned lawnmower patterns that provide dense, overlapping coverage over reference grids, using a down-looking stereo camera rig with LED illumination for high-resolution imaging. Its navigation system fuses~\gls{DVL}, attitude, depth, acoustic positioning, and visual~\gls{SLAM}~\citep{mahon_efficient_2008} to produce self-consistent trajectories and precisely georeferenced imagery suitable for 3D reconstruction and repeat surveys of permanent reference sites.

\begin{figure}[!ht]
    \centering
    \includegraphics[width=\textwidth]{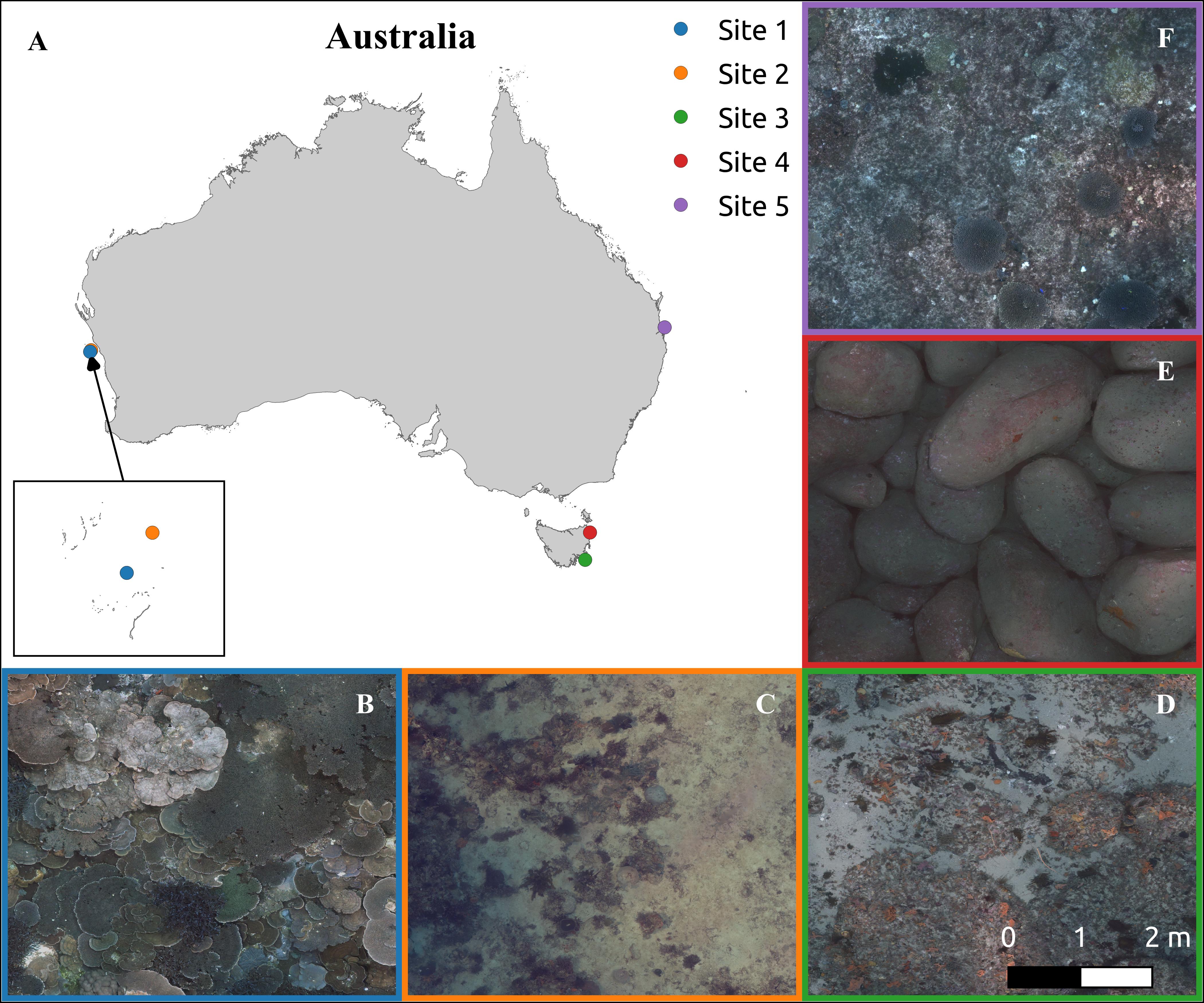}
    \caption{
        Map with overview of the benthic reference sites in our dataset. \textbf{(A)} Map of Australia and the geographic location of each site. \textbf{(B-F)} Orthomosaics showing seafloor patches with characteristics representative for the following benthic reference sites; \textbf{(B)} Site 1, \textbf{(C)} Site 2, \textbf{(D)} Site 3, \textbf{(E)} Site 4, and \textbf{(F)} Site 5. The orthomosaic patches are rendered at the same spatial scale, indicated by the scalebar in the lower right corner. Base map data 
        \textcopyright{} Commonwealth of Australia (Australian Bureau of Statistics) 2021, 
        ASGS Edition~3 digital boundaries, used under CC~BY~4.0~\citep{australian_bureau_of_statistics_australian_2021}.
    }
    \label{fig:reference_site_map}
\end{figure}

The benchmark dataset is derived from the archive of benthic imagery collected by AUV Sirius across multiple~\gls{IMOS} campaigns and deployments and made accessible through the IMOS–UMI Squidle+ web portal~\citep{friedman_squidle_2025}. From this archive, we sought regions that (i) were mapped with dense, overlapping coverage during individual visits and (ii) were revisited at least three times, providing overlapping views of the same seafloor patch over time. To automatically detect such regions, the clustering-based method of~\cite{larsen_geometric_2023} was applied to the georeferenced image locations on Squidle+, yielding spatial clusters that correspond to densely mapped areas and serve as candidate benthic reference sites. The raw data for deployments containing these clusters was then retrieved from internal~\gls{ACFR} data servers. From the set of candidate clusters, five reference sites were selected to constitute the final benchmark dataset. Each site required at least three visits from different years with overlapping coverage, sufficient image counts and georeferencing quality to support robust 3D reconstruction. Sites dominated by highly dynamic cover such as kelp were excluded due to reconstruction quality. The selected sites span depths of approximately 18–45~m and encompass diverse habitat types, including sparse and dense coral reefs, soft sediment bottom, rock reefs, and boulder reefs. \Cref{fig:reference_site_map} summarizes the spatial distribution and representative orthomosaic patches for each site, and \Cref{tab:site_overview} provides an overview of their key properties.

\begin{table}[!ht]
    \centering
    \footnotesize 
    \begin{tabular}{cccccp{0.35\textwidth}}
        \toprule
        Site & Geohash & Visit Years & Depth range & Image Count 
            & Site Description \\
        \midrule
        - & - & - & m & - & - \\
        \midrule
        Site 1 & QDCH0FTQ & 2010, 2011, 2012, 2013 & 17 - 20 & 17794
            & Dense coral reef \\
        Site 2 & QDCHDMY1 & 2011, 2012, 2013, 2017 & 27 - 35 & 20944
            & Dense coral reef and soft sediment \\
        Site 3 & R23685BC & 2010, 2012, 2014 & 40 - 43 & 15146
            & Rock reef with corals and sponges and sandy substrate \\
        Site 4 & R29MRD5H & 2009, 2011, 2013 & 25 - 36 & 21362
            & Boulder reef \\
        Site 5 & R7JJSKXQ & 2010, 2012, 2013 & 18 - 21 & 19444
            & Rock reef with corals and sponges \\
        \bottomrule 
    \end{tabular}
    \caption{Overview of key properties for the benthic reference sites in our dataset.}
    \label{tab:site_overview}
\end{table}

\subsection{Image Color Correction}
\label{sec:method_image_color_correction}

\begin{figure}[!ht]
    \centering
    \includegraphics[width=\textwidth]{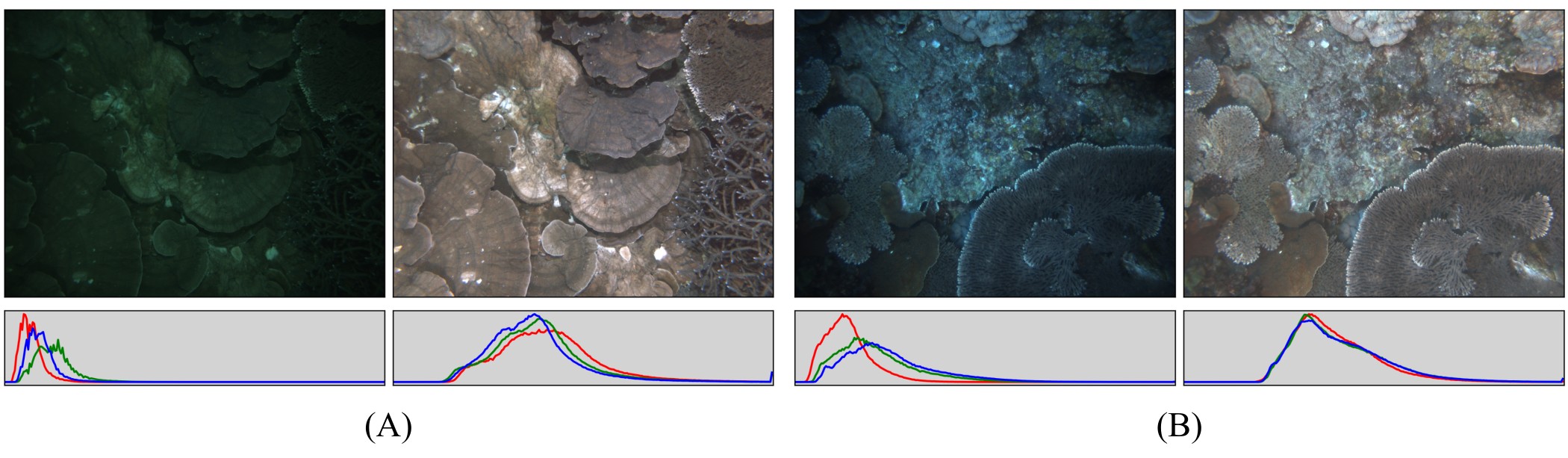}
    \caption{
        (A) Raw (left) and color corrected (right) image from the 2010 visit to Site 1.
        (B) Raw (left) and color corrected (right) image from the 2013 visit to Site 1.
    }
    \label{fig:color_correction_examples}
\end{figure}

Imagery captured by underwater robotic platforms is affected by wavelength-dependent attenuation, backscatter, and non-uniform artificial illumination, which together cause strong color distortions and spatial brightness variations. To improve the color consistency of our dataset, we apply the multi-image gray-world algorithm~\citep{bryson_colour-consistent_2013}, which is designed to correct spatial color variations caused by lens vignetting and non-uniform lighting. In contrast to the conventional gray-world assumption over a single image~\citep{buchsbaum_spatial_1980}, this method estimates per-pixel statistics over multiple images from the same camera, and then applies a linear per-pixel correction that equalizes the mean and variance of the color channels across the image dataset. For each camera and visit, we estimate these correction factors from the raw images and apply the resulting linear transform to all images for that camera and visit, yielding color-corrected imagery with more uniform illumination and improved color balance. Images are normalized to the $\left[0, 1 \right]$ intensity range before correction, and we set the desired mean and standard deviation of the corrected intensities to $\imageDesiredMean \left( \imageColorChannel \right) = 0.35$ and $\imageDesiredStd(\imageColorChannel)=0.12$, respectively. These values were chosen empirically to produce moderate brightness and contrast across sites while limiting saturation of bright pixels from bleached corals or highly reflective substrates. Figure~\ref{fig:color_correction_examples} shows examples of raw and color-corrected images from two visits. In addition to the corrected imagery, we also provide the original raw images in the dataset so that users can apply alternative color correction or enhancement methods if desired.

\subsection{Geometric Reconstruction and Registration}
\label{sec:method_geometric_processing}

\begin{figure}[!ht]
    \centering
    \includegraphics[width=\textwidth]{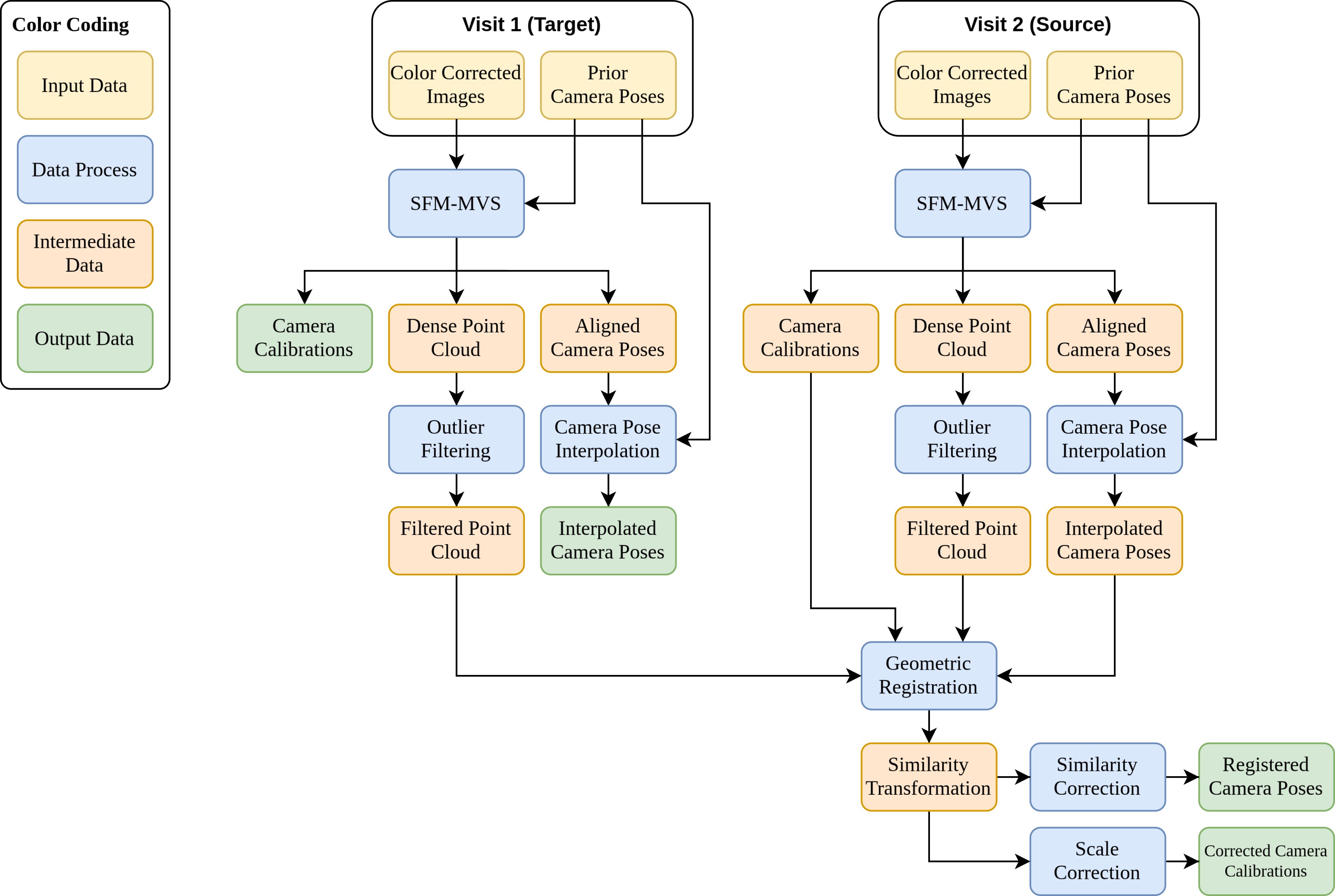}
    \caption{
        Single line diagram of the geometric reconstruction and registration workflow for an example with two visits. In the example Visit 1 is the registration target, while Visit 2 is the registration source being registered to Visit 1. Note that for the target visit, the~\gls{SFM}-estimated camera calibrations and interpolated camera poses are the final calibrations and camera poses, while for the source visit they are corrected by the similarity transformation estimated by the geometric registration.
    }
    \label{fig:geometric_processing_workflow}
\end{figure}

To register camera poses across visits to benthic reference sites, we apply a geometric approach based on 3D reconstruction and registration, similar to previous work \citep{larsen_geometric_2023, boittiaux_eiffel_2023}. \Cref{fig:geometric_processing_workflow} outlines the workflow. We estimate camera calibrations, poses, and scene geometry using \gls{SFM} and \gls{MVS}. Color-corrected stereo image pairs (\Cref{sec:method_image_color_correction}) and prior poses from a \gls{SLAM} system \citep{mahon_efficient_2008} serve as inputs for each visit. \gls{SFM} and \gls{MVS} is performed in Agisoft Metashape Professional \citep{agisoft_llc_agisoft_2022}, producing globally aligned camera poses, intrinsic and extrinsic calibrations, and dense point clouds. When \gls{SFM} fails on certain trajectory segments, typically due to poor image overlap or dynamic content, we compute a local similarity transform $\mathbf{S}_{s} \in \lieGroupSim{3}$ using Umeyama’s least-squares method \citep{umeyama_least-squares_1991}. The transform aligns prior poses with neighboring \gls{SFM} estimates, yielding locally corrected poses. The union of these locally aligned and \gls{SFM}-aligned poses are referred to as interpolated camera poses. Once the camera poses for each visit are estimated, we perform geometric registration between the visits to express all camera poses in the same reference frame. The geometric registration is done using the dense 3D point clouds derived from~\gls{MVS}. We process the dense point clouds by discarding 3D points observed by fewer than three stereo image pairs, which removes many noisy or weakly constrained points arising from~\gls{MVS}. For each site, we manually select a reference visit (target) as the visit with the greatest spatial overlap with the remaining visits; all other visits (sources) for that site are registered to this reference. For each source visit, we estimate a 3D similarity transform that we apply to all camera poses of that visit. We also correct the scale of the extrinsic calibration of the stereo rig.

For each site, we define a four-stage registration pipeline. At each stage, we downsample the outlier-filtered point clouds to a prescribed spatial resolution using voxel downsampling and estimate a rigid or similarity transformation between the source and target point clouds. In Stage 1, we perform global registration using~\gls{FPFH} descriptors~\citep{rusu_fast_2009} and a~\gls{RANSAC}-based estimator~\citep{fischler_random_1981} to obtain an initial similarity transform between the source and target visits. In Stages 2–4, we fix the scale parameter and refine the rotation and translation using colored~\gls{ICP}~\citep{park_colored_2017} at progressively finer spatial resolutions, yielding the final similarity transform for that source visit. The final similarity transform is used to update the camera poses and extrinsic calibration for the source visit. This process is repeated for all source visits for a given site. The registration pipeline is implemented using Open3D~\citep{zhou_open3d_2018}, which provides~\gls{FPFH}, \gls{RANSAC}-based global registration, and colored~\gls{ICP} routines. The configurations of the registration pipeline are provided in the Supplementary Materials . To quantify the registration error between visits, we leverage the Euclidean distance between corresponding points:

\begin{equation}
    \label{eq:geometric_registration_distance}
    \mathbf{e}_{ij} = \norm{\mathbf{p}_{i} - \mathbf{p}_{j}}^{2},
    \quad \left( \mathbf{p}_{i}, \mathbf{p}_{j} \right) \in \mathcal{A},
\end{equation}

where $\mathbf{p}_{i}$ and $\mathbf{p}_{j}$ are corresponding 3D points from the target and source visit, respectively, and $\mathcal{A}$ is the set of point cloud correspondences from the final colored \gls{ICP} stage.

\subsection{Estimating Image Seafloor Footprints}
\label{sec:method_image_footprint_estimation}

Estimating image footprints on the seafloor is central to how visual localizations are classified as true or false in this work. Rather than relying on fixed location distance thresholds, which do not account for variations in vehicle altitude and seafloor relief, the proposed approach uses image-footprint overlap to decide whether two images actually observe part of the same seafloor patch. Similar footprint-based reasoning has previously been used in visually augmented navigation~\citep{eustice_visually_2008} and view-based~\gls{SLAM}~\citep{mahon_efficient_2008} to propose image-to-image links and loop-closure hypotheses for~\glspl{AUV}. Here, this idea is adapted for long-term \gls{VPR} by combining calibrated cameras with range estimates to obtain per‑view (per‑image) 3D footprints. The top-down overlap of these footprints then provides a content-based criterion for defining true localizations.

In this work, we use the following coordinate reference frames:
\begin{itemize}
    \item The global frame is the WGS84 Earth‑fixed frame, in which positions are represented by geodetic latitude, longitude, and ellipsoidal height. A 3D point in the global frame is denoted $\indexed{\objectCoordVec}{g}$.
    \item The local frame is the 3D Cartesian frame fixed to the environment following North--East--Down (NED) convention. It is defined with respect to a reference point on WGS84, enabling conversion between local coordinates $\indexed{\objectCoordVec}{l}$ and global coordinates $\indexed{\objectCoordVec}{g}$.
    \item The vehicle frame is the 3D frame fixed to the vehicle following the SNAME convention~\citep{fossen_handbook_2011}, with axes x positive forward, y positive to starboard, and z downward. The origin of the vehicle frame is chosen at the optical center of the left camera of the stereo rig.
\end{itemize}

Each camera is modeled as an ideal pinhole camera without skew with intrinsics $\intrinsicMatrix$ \citep{hartley_multiple_2003}, and we adopt the OpenCV camera and pixel coordinate conventions (x right, y down, z along the optical axis; pixel coordinates with origin in the top-left corner). A 3D point in the frame of camera view $c_{k}$ at time $k$ is denoted as $\indexed{\objectCoordVec}{c_{k}}$.


\subsubsection{Camera Model}
\label{sec:method_camera_model}

Under this model, the calibrated intrinsic matrix is

\begin{equation}
    \label{eq:camera_intrinsic_matrix}
    \intrinsicMatrix = \begin{bmatrix}
        \focalLengthX & 0 & \principalPointX \\
        0 & \focalLengthY & \principalPointY \\
        0 & 0 & 1
    \end{bmatrix},
\end{equation}

where $\focalLengthX$ and $\focalLengthY$ are the focal lengths in $x$ and $y$ direction, respectively, and $\left[ \principalPointX, \principalPointY \right]^{\transpose}$ is the principal point of the camera measured in pixels. The projective mapping from a 3D point in the local frame to a 2D point in the pixel frame of camera view $c_{k}$ is given as:

\begin{equation}
    \label{eq:camera_projective_mapping}
    \pixelCoordVecHomo_{j}
        = 
        \begin{bmatrix}
            \intrinsicMatrix_{c_{k}} & \mathbf{0} \\
        \end{bmatrix}
        \indexed{\objectCoordVecHomo}{c_{k}}_{j}
        =
        \begin{bmatrix}
            \intrinsicMatrix_{c_{k}} & \mathbf{0} \\
        \end{bmatrix}
        \transformRigid{l}{c_{k}} \indexed{\objectCoordVecHomo}{l}_{j}
\end{equation}

Here $\pixelCoordVec_{j} = \left[ \pixelCoordX_{j}, \pixelCoordY_{j} \right]^{\transpose}$ is a point $j$ in the pixel frame of camera view $c_{k}$, and $\pixelCoordVecHomo_{j} = \left[ \pixelCoordVec_{j}^{\transpose}, 1 \right]^{\transpose}$ is its homogeneous representation. $\intrinsicMatrix_{c_{k}}$ is the intrinsic matrix for the camera of view $c_{k}$. $\indexed{\objectCoordVec}{l}_{j} = \left[ \indexed{x}{l}_{j}, \indexed{y}{l}_{j}, \indexed{z}{l}_{j} \right]^{\transpose}$ is a 3D point expressed in the local frame, and $\indexed{\objectCoordVecHomo}{l}_{j} = \left[ \indexed{\objectCoordVec}{l}_{j}^{\transpose}, 1 \right]^{\transpose}$ is its homogeneous representation. The transformation $\transformRigid{l}{c_{k}} = \transformRigid{v}{c} \transformRigid{l}{v_{k}} \in \lieGroupSE{3}$ maps points from the local frame to the frame of camera view $c_{k}$, combining the vehicle pose $\transformRigid{l}{v_{k}}$ at time $k$ in the local frame with the time‑invariant extrinsics $\transformRigid{v}{c}$ of the camera $c$ relative to the vehicle. For a calibrated camera, we can invert the projective mapping in~\Cref{eq:camera_projective_mapping}, and estimate the 3D point on the seafloor corresponding to the image pixel $\pixelCoordVec_{j}$ with the following equations:

\begin{subequations}
\begin{equation}
    \label{eq:inverse_projective_mapping}
    \indexed{\objectCoordVec}{c_{k}}_{j} 
        = \range_{j} \cdot \intrinsicMatrix_{c_{k}}^{-1} \pixelCoordVecHomo_{j}
\end{equation}
\begin{equation}
    \label{eq:transform_point_camera_to_local}
    \indexed{\objectCoordVecHomo}{l}_{j}
        = \transformRigid{c_{k}}{l} \indexed{\objectCoordVecHomo}{c_{k}}_{j}
\end{equation}
\end{subequations}

Here $\intrinsicMatrix_{c_{k}}^{-1} \pixelCoordVecHomo_{j}$ gives the ray direction up to scale in the camera frame, which is then scaled by the range $\range_{j}$. The range is defined as the distance from the optical center $\cameraOriginVec_{c_{k}}$ of camera view $c_{k}$ to the seafloor along the optical (z-)axis for pixel $j$. $\transformRigid{c_{k}}{l} \in \lieGroupSE{3}$ is the rigid-body transformation from the frame of camera view $c_{k}$ to the local frame.


\subsubsection{Range Map Fusion}
\label{sec:method_range_map_fusion}

\newcommand{\rangeSampleStereo}{z^{stereo}}
\newcommand{\rangeSampleRel}{z^{rel}}
\newcommand{\rangeSampleFused}{z^{fused}}

Accurate footprint estimation requires spatially consistent range information across the field of view, particularly at image borders where footprint vertices are defined. However, \gls{MVS} range maps (\Cref{sec:method_geometric_processing}) are often incomplete, so valid values are frequently missing near image boundaries. In addition, camera views that are unaligned by \gls{SFM} lack \gls{MVS}-derived range maps, making them unsuitable for footprint estimation. To obtain metric range estimates without relying on multi‑view overlap, we compute stereo‑derived range maps using the calibrated stereo rig and HitNet~\citep{tankovich_hitnet_2021} for left–right disparity. Stereo reconstruction provides metrically accurate range for individual image pairs and thus supports all camera views, including those not aligned by \gls{SFM}, but often misses fine structural details and shows geometric inconsistencies where left–right overlap is limited. To compensate for these shortcomings, we fuse the stereo‑derived metric range maps with monocular relative range maps from Depth~Anything~V2~\citep{yang_depth_2024}, which provide complete but scale‑ambiguous predictions. This fusion combines stereo metric accuracy with monocular spatial completeness, yielding dense, metrically consistent range maps for all camera views.

Following recent work on aligning monocular range maps with a global scale and offset~\citep{ganj_hybriddepth_2025, zhang_spade_2025}, we adopt a similar strategy to fuse the stereo and monocular estimates. For each image, the stereo range map is first reprojected so that both maps share the same (distorted) pixel frame, and stereo values outside $[0.2,6.0]$~m are masked out to remove close‑range artifacts and far‑range noise; the same mask is applied to the monocular map. From the remaining pixels, 10\% are randomly sampled to form pairs of stereo and monocular ranges, and a global scale and offset $\left(a, b\right)$ are estimated by robust linear regression with a Huber loss:

\begin{equation}
    \label{eq:range_fusion_optimization}
    \underset{a, b}{\text{argmin}}\;
    \sum_{i} \rho \left( 
        \norm{\rangeSampleFused_{i} - \rangeSampleStereo_{i}}^{2} 
    \right),
\end{equation}

where the fused range for each pixel $i$ is defined as

\begin{equation}
    \label{eq:range_fusion_fused_range}
    \rangeSampleFused_{i} = a \cdot \rangeSampleRel_{i} + b.
\end{equation}

\begin{figure}[!ht]
    \centering
    \includegraphics[width=\textwidth]{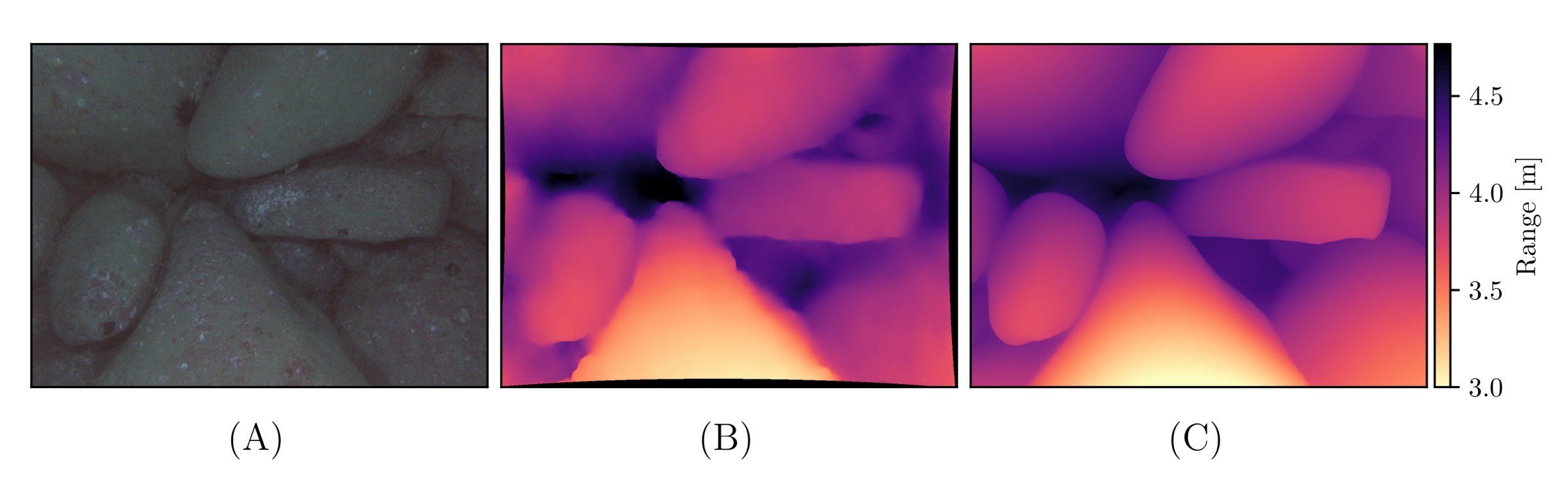}
    \caption{
        An example showing the fusion of relative and metric stereo-derived range maps for an image from 2009 visit to Site 4. (A) The color corrected image, (B) the stereo range map reprojected into the pixel coordinate system of the original image, and (C) the range map estimated by fusing the relative range map with the stereo-derived range map. The stereo-derived and fused range maps are rendered with the same color map to indicate the range. Overall, the fused range maps show sharper edges and do not exhibit the geometric inconsistencies on the left side seen in the stereo range map, which are believed to be caused by limited left–right image overlap.
    }
    \label{fig:range_map_fusion_example}
\end{figure}

The resulting parameters $\left(a, b\right)$ are applied to the full‑resolution monocular map to obtain a dense, metrically scaled range map, which is then used for footprint estimation. \Cref{fig:range_map_fusion_example} shows an example with the stereo range map and fused range map for an image captured by the left (RGB) camera of the stereo rig.

\subsubsection{Image Footprint and Overlap Estimation}
\label{sec:method_image_footprint_and_overlap_estimation}

In our approach, we estimate the image footprints by using the inverse projective mapping in \Cref{eq:inverse_projective_mapping} for the corners of each image. We estimate the range $\range_{j}$ for each corner $j$ as the median range in a $30 \times 30$ pixel patch in the corresponding corner of the fused range maps from \Cref{sec:method_range_map_fusion}. Using \Cref{eq:inverse_projective_mapping} with the pixel coordinates $\pixelCoordVec_{j}$ and range estimate $\range_{j}$ for each image corner, and transforming the 3D points using \Cref{eq:transform_point_camera_to_local}, we obtain the set of 3D points (3D polygon) in the local frame $\indexed{\footprintSet_{c_{k}}}{l}$ representing the image footprint for camera view $c_{k}$: 

\begin{equation}
    \label{eq:image_footprint_local}
    \indexed{\footprintSet_{c_{k}}}{l}
     = \left \{
        \indexed{\objectCoordVec_{c_{k}, 1}}{l},
        \indexed{\objectCoordVec_{c_{k}, 2}}{l},
        \indexed{\objectCoordVec_{c_{k}, 3}}{l},
        \indexed{\objectCoordVec_{c_{k}, 4}}{l}
    \right \}
\end{equation}

To perform spatial operations on the image footprints, we georeference them by converting the footprints expressed in the local NED frame, $\indexed{\footprintSet_{c_{k}}}{l}$, to footprints expressed in global WGS84 frame, $\indexed{\footprintSet_{c_{k}}}{g}$. This is primarily a choice of convenience as we no longer need to keep track of the reference point for the local frame. Subsequently, all spatial operations are performed on the footprints in the global frame. In order to perform spatial operations between image footprints, the 3D footprints are converted to 2D by discarding the ellipsoidal height. This effectively projects all footprints onto a single planar reference surface, an acceptable approximation when elevation variations are small relative to the spatial extent of the site. In our dataset, the maximum elevation difference within a reference site is \textasciitilde9 m, and neighboring footprints typically differ by only a few meters in height at maximum. Under these conditions, the 2D projection introduces negligible spatial error. However, the approximation becomes unreliable in areas with pronounced vertical structure, such as underwater cliffs, steep walls, or caves, where the surfaces of interest vary primarily along the vertical axis and cannot be represented as lying on a common horizontal plane. Let $\indexed{\footprintSetProj}{g}$ denote the 2D conversion of the 3D footprint $\indexed{\footprintSet}{g}$ in the global frame. For the 2D footprints $\indexed{\footprintSetProj_{c_{k}}}{g}$ and $\indexed{\footprintSetProj_{c_{j}}}{g}$ of camera view $c_{k}$ and $c_{j}$, respectively, we denote the spatial intersection and spatial union operators as:

\begin{subequations}
\begin{equation}
    \label{eq:footprint_spatial_intersection}
    \indexed{\footprintSetProj_{c_{k}}}{g} \spatialIntersect 
    \indexed{\footprintSetProj_{c_{j}}}{g} = 
        \SpatialIntersectFun \left( 
            \indexed{\footprintSetProj_{c_{k}}}{g},
            \indexed{\footprintSetProj_{c_{j}}}{g}
        \right)
\end{equation}
\begin{equation}
    \label{eq:footprint_spatial_union}
    \indexed{\footprintSetProj_{c_{k}}}{g} \spatialUnion 
    \indexed{\footprintSetProj_{c_{j}}}{g} = 
        \SpatialUnionFun \left( 
            \indexed{\footprintSetProj_{c_{k}}}{g},
            \indexed{\footprintSetProj_{c_{j}}}{g}
        \right)
\end{equation}
\end{subequations}

For non-overlapping footprints $\SpatialIntersectFun \left( \cdot \right)$ and $\SpatialUnionFun \left( \cdot \right)$ return empty sets. To detect if camera view $c_{k}$ and $c_{j}$ have overlapping footprints, we check that the spatial intersection between their 2D footprints, $\indexed{\footprintSetProj_{c_{k}}}{g}$ and $\indexed{\footprintSetProj_{c_{j}}}{g}$, respectively, is not an empty set, i.e.:

\begin{equation}
    \label{eq:footprint_overlap_criterion}
    \indexed{\footprintSetProj_{c_{k}}}{g} \spatialIntersect 
    \indexed{\footprintSetProj_{c_{j}}}{g} \neq \emptyset
\end{equation}

In order to quantify the amount of overlap of image footprints relative to the size of the footprints, we define the footprint \gls{IOU} as:

\begin{equation}
    \label{eq:footprint_iou}
    \footprintIou_{kj} = 
        \frac{
            \SpatialAreaFun \left(
                \indexed{\footprintSetProj_{c_{k}}}{g} \spatialIntersect 
                \indexed{\footprintSetProj_{c_{j}}}{g} 
            \right)
        }{
            \SpatialAreaFun \left(
                \indexed{\footprintSetProj_{c_{k}}}{g} \spatialUnion 
                \indexed{\footprintSetProj_{c_{j}}}{g}
            \right)
        }
\end{equation}

\begin{figure}[!ht]
    \centering
    \includegraphics[width=\textwidth]{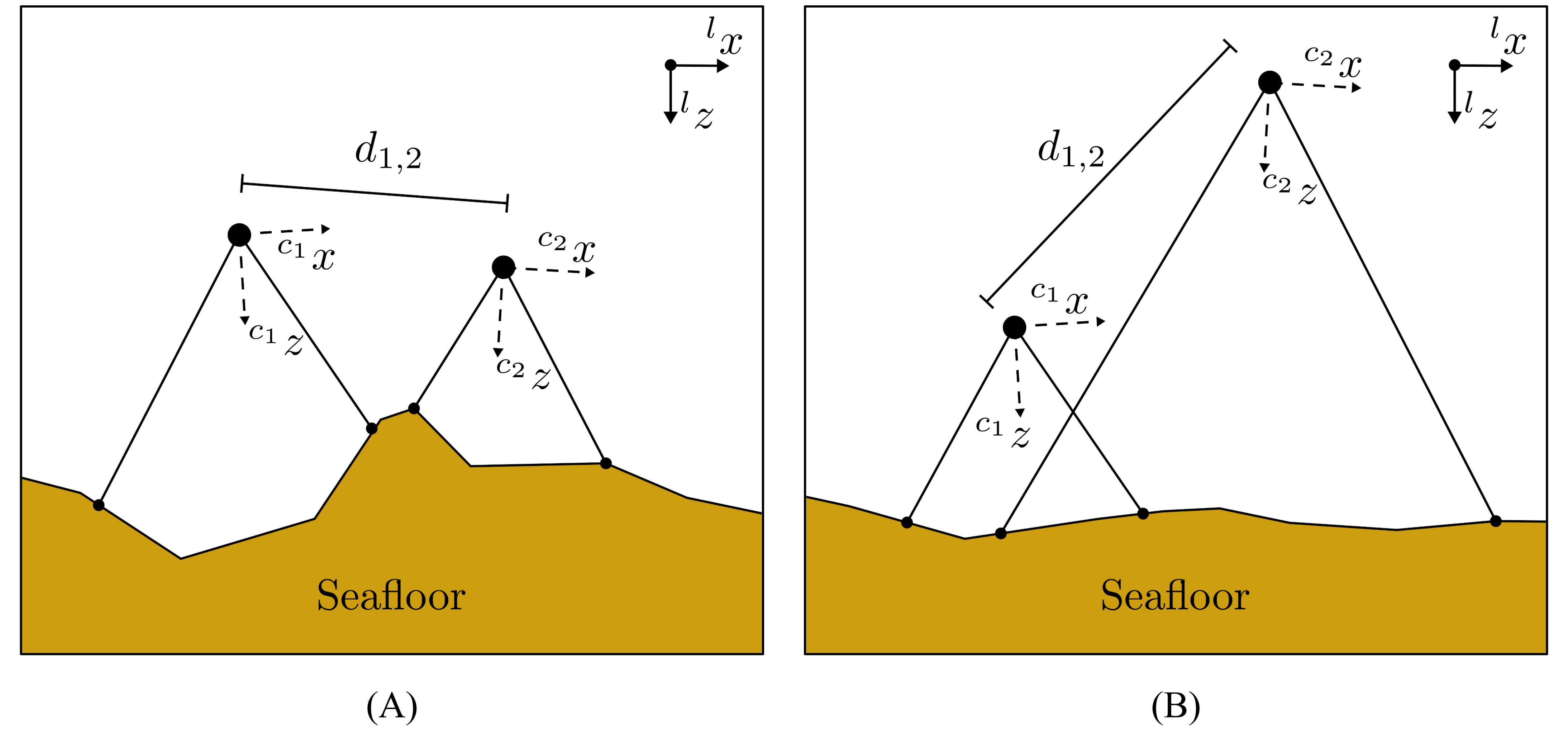}
    \caption{
        Illustration of a 2D simplified model of our footprint estimation method for two scenarios. Panel (A) shows a scenario where local terrain relief causes non-overlapping image footprints for two spatially close camera views. Panel (B) shows a scenario where large altitude differences causes spatially distant camera views to have overlapping image footprints.
    }
    \label{fig:method_footprint_estimation_illustration}
\end{figure}

Here $\SpatialAreaFun \left( \cdot \right)$ is the function that returns the spatial area of a 2D polygon. The spatial intersect, union, and area functions, $\SpatialIntersectFun \left( \cdot \right)$, $\SpatialUnionFun \left( \cdot \right)$, and $\SpatialAreaFun \left( \cdot \right)$, respectively, are implemented with Shapely~\citep{gillies_shapely_2023} which uses GEOS~\citep{geos_contributors_geos_2025} in its backend to perform geometric operations. \Cref{fig:method_footprint_estimation_illustration} shows a simplified 2D model of our footprint estimation method for two scenarios. Panel (A) shows a scenario where local terrain relief causes non-overlapping image footprints for two spatially close camera views. Panel (B) shows a scenario where large altitude differences causes spatially distant camera views to have overlapping image footprints.

\newcommand{\visitSet}{\mathcal{V}}
\newcommand{\cameraViewSet}{\mathcal{C}}
\newcommand{\cameraViewLinkSet}{\mathcal{L}}
\newcommand{\descriptorVec}{\mathbf{e}}
\newcommand{\descriptorSet}{\mathcal{E}}
\newcommand{\footprintIouThreshold}{\tau_{f}}
\newcommand{\distanceThreshold}{\tau_{d}}


\newcommand{\countTPK}{\mathrm{TP@K}}
\newcommand{\countFPK}{\mathrm{FP@K}}
\newcommand{\countFNK}{\mathrm{FN@K}}
\newcommand{\metricInfoRecallK}{\mathrm{IRRecall@K}}
\newcommand{\metricVprRecallK}{\mathrm{Recall@K}}
\newcommand{\rankCutoff}{K}

\subsection{Evaluating Long-Term Visual Place Recognition}
\label{sec:method_evaluating_long_term_vpr}

In this paper, we treat long-term~\gls{VPR} as an image retrieval problem: each image is mapped into a vector representation $\descriptorVec \in \mathbb{R}^{d}$, known as an image descriptor or embedding, and image-level nearest-neighbor search in descriptor space is used to retrieve candidate images from a database of images with associated location estimates.

\subsubsection{Selected Visual Place Recognition Models}
\label{sec:method_selected_vpr_models}

To benchmark long-term~\gls{VPR} on our dataset, we evaluate eight representative global-descriptor models spanning both~\gls{CNN}- and~\gls{VIT}-based architectures. The~\gls{CNN}-based methods include NetVLAD~\citep{arandjelovic_netvlad_2016}, MixVPR~\citep{ali-bey_mixvpr_2023}, CosPlace~\citep{berton_rethinking_2022}, and EigenPlaces, which represent widely used global descriptor designs for place recognition. The~\gls{VIT}-based methods build on DINOv2~\citep{oquab_dinov2_2024} features and include AnyLoc~\citep{keetha_anyloc_2024}, CliqueMining~\citep{izquierdo_close_2025}, SALAD~\citep{izquierdo_optimal_2024}, and MegaLoc~\citep{berton_megaloc_2025}, providing a diverse set of transformer-based global descriptors for long-term~\gls{VPR}.

\subsubsection{Evaluation Protocol and Metrics}
\label{sec:method_evaluation_protocol_and_metrics}

We first specify the input modality and preprocessing used for all \gls{VPR} models. For benchmarking, we use only the RGB images from the left camera of the stereo rig, color‑corrected with the methodology outlined in \Cref{sec:method_image_color_correction}. This matches the input configuration used to train and evaluate all compared methods, NetVLAD, MixVPR, CosPlace, EigenPlaces, AnyLoc, CliqueMining, SALAD, and MegaLoc, which are designed for single‑view RGB imagery and for which published results are reported under this setting. This choice isolates long‑term appearance and structural changes, without introducing additional confounding factors from sensor‑modality fusion or model re‑training.

Since the sites in our dataset exhibit different seafloor characteristics, we evaluate the~\gls{VPR} models on a per-site basis. For each site, we split the visits into pairs $\left( \visitSet_{Q}, \visitSet_{D} \right)$, where $\visitSet_{Q}$ represents a new unseen visit and $\visitSet_{D}$ represents a previously seen visit. We refer to $\visitSet_{Q}$ and $\visitSet_{D}$ as the query and database visit to reflect their role when performing image retrieval. We use a temporal pairing strategy for the visits, where a query visit is only paired with visits that where conducted before it in time. For each visit pair, we denote the set of camera views from the query and database visit as $\cameraViewSet_{Q}$ and $\cameraViewSet_{D}$, respectively. For every camera view in the query visit, we create links to the camera views in the database visit by using the georeferenced 3D image footprints estimated with the methodology in \Cref{sec:method_image_footprint_and_overlap_estimation}. These camera view links composes our ground truth, i.e. what we consider to be true localizations. We use the criterion in \Cref{eq:footprint_overlap_criterion} to find pairs of query and database camera views with overlapping image footprints. For query and database camera views with overlapping footprints, we create a set of camera view links, denoted $\cameraViewLinkSet_{Q,D}$ by setting a lower threshold $\footprintIouThreshold$ for the footprint~\gls{IOU} defined in \Cref{eq:footprint_iou}. The set of camera view links is defined as:

\begin{equation}
    \label{eq:camera_footprint_linking}
    \cameraViewLinkSet_{Q,D} = \left\{
        \left( c_{q}, c_{d} \right) 
        \mid \footprintIou_{qd} > \footprintIouThreshold,
        \; c_{q} \in \cameraViewSet_{Q},
        \; c_{d} \in \cameraViewSet_{D}
    \right\}
\end{equation}

To design a conservative lower footprint \gls{IOU} threshold $\footprintIouThreshold$ that makes us confident that two camera views have overlapping footprints, we consider a simplified scenario where both cameras observe a flat seafloor patch from the same altitude while looking straight down, with their footprints aligned along the largest side. Under the assumption that the registration error is a purely horizontal translation, normal to the aligned footprint side and smaller than the smallest footprint side $L_{f}$, the induced footprint \gls{IOU} is

\begin{equation}
    \label{eq:lower_footprint_iou_estimate}
    \footprintIou_{e} 
        = \frac{
            W_{f} \cdot t_{e}
        }{
            2 \cdot W_{f} \cdot L_{f} - W_{f} \cdot t_{e}
        }
        = \frac{
            t_{e}
        }{
            4 \cdot a \cdot \tan \left( 0.5 \cdot \fieldOfView \right) - t_{e}
        },
\end{equation}

where $t_{e}$ is the horizontal translation and $a$ is the altitude. $W_{f}$ and $L_{f}$ are the lengths of the largest and smallest footprint sides, respectively. $\fieldOfView$ is the field of view along the axis corresponding to $L_{f}$. With this simplified model, choosing $\footprintIouThreshold=\footprintIou_{e}$ makes us confident that the two footprints overlap for registration errors up to $t_{e}$.


For each of the RGB images associated with the camera views we use the~\gls{VPR} model to create the set of image descriptors $\descriptorSet_{Q}$ and $\descriptorSet_{D}$ for the query and database camera views, respectively. For every query camera view $c_{q}$, we perform a similarity search between the image descriptor of the query camera view, $\descriptorVec_{q}$, and the set of image descriptors for the database camera views, $\descriptorSet_{D}$. The similarity search finds the $\rankCutoff$ candidates from the database image descriptors that minimize the L2 distance to the image descriptor of the query camera view $\descriptorVec_{q}$. The similarity search is defined as:

\begin{equation}
    \label{eq:image_similarity_search}
    \cameraViewSet_{q,p}^{(\rankCutoff)}
        = \left\{ c_{p}^{(1)}, \dots, c_{p}^{(\rankCutoff)} \right\}
        = \underset{
            c_{p} \in \cameraViewSet_{D}
        }{
            \mathrm{TopK}
        } \left(
            - \norm{ \descriptorVec_{q} - \descriptorVec_{p}}^{2}
        \right)
\end{equation}

Here $\cameraViewSet_{q,p}^{(\rankCutoff)}$ is the ordered set of candidate database camera views for query camera view $c_{q} \in \cameraViewSet_{Q}$, $\descriptorVec_{q} \in \descriptorSet_{Q}$ is the image descriptor of the query camera view, and $\descriptorVec_{p} \in \descriptorSet_{D}$ is the image descriptor of the candidate database camera view $c_{p} \in \cameraViewSet_{D}$. The $k$-th nearest candidate database camera view is denoted $c_{p}^{(k)}$. We refer to $k$ as the retrieval or candidate rank, and $\rankCutoff$ as the rank cutoff. In our implementation, we implement $\topKFun \left( \cdot \right)$ using FAISS~\citep{douze_faiss_2025}.

For each visit pair, we evaluate localizations for each query camera view $c_{q}$ using the set of retrieved database camera views, $\cameraViewSet_{q,p}$, and the set of footprint-based ground-truth links between the query and database camera views, $\cameraViewLinkSet_{Q,D}$. We consider a \gls{VPR} candidate as a true positive if the pair of query and database camera views $\left( c_{q}, c_{p} \right) \in \cameraViewSet_{q,p}^{(K)}$ matches the ground-truth links, i.e. $\left( c_{q}, c_{p} \right) \in \cameraViewLinkSet_{Q,D}$. Consequently, a pair of query and database camera view in the ground-truth links is considered a false negative if it is not in 
the \gls{VPR} retrieved set, i.e. $\left( c_{q}, c_{d} \right) \notin \cameraViewSet_{q,p}^{(K)}$. We adopt the \gls{VPR}‑specific definition of $\mathrm{Recall@K}$ to ensure comparability with standard \gls{VPR} benchmarks, where this metric is the prevailing evaluation metric. $\metricVprRecallK$ directly quantifies the proportion of query images that have at least one correct match within the top‑K retrieved candidates, which provides an intuitive measure of localization success and corresponds to what is typically referred to as $\mathrm{HitRate@K}$ in information retrieval. In typical robotic localization pipelines, \gls{VPR} operates as a retrieval module whose top‑ranked candidates are subsequently validated using geometric verification (for example, image keypoint‑based pose estimation). Consequently, the standard way to assess \gls{VPR} performance is to measure how often the module returns at least one correct candidate within the top‑K results for each query, as captured by $\metricVprRecallK$. Formally, we define $\metricVprRecallK$ as

\begin{equation}
    \label{eq:vpr_recall@K}
    \metricVprRecallK
        = \frac{
            \text{Number of queries with at least one correct candidate}
        }{
            \text{Number of valid queries}
        }
\end{equation}

Here, a query is considered recognized if any of the retrieved pairs $\left( c_{q}, c_{p} \right) \in \cameraViewSet_{q,p}^{(K)}$ is a true positive, i.e. $\left( c_{q}, c_{p} \right) \in \cameraViewLinkSet_{Q,D}$. We consider unlinked query camera views, i.e. query camera views that do not have any database camera views with overlapping footprints, as invalid queries. Consequently, we do not consider the \gls{VPR} retrievals for those queries when calculating $\metricVprRecallK$. While $\metricVprRecallK$ evaluates recognition success on a per-query basis, $\metricInfoRecallK$ follows the classical formulation from information retrieval, measuring the proportion of all relevant query–database pairs that are correctly retrieved. We use $\metricInfoRecallK$ specifically when studying how different definitions of ground truth affect retrieval performance. The information retrieval recall $\metricInfoRecallK$ is defined as

\begin{equation}
    \label{eq:information_recall@K}
    \metricInfoRecallK = \frac{\countTPK}{\countTPK + \countFNK},
\end{equation}

where $\countTPK$ and $\countFNK$ denote, respectively, the number of true‑positive and false‑negative query–database pairs within the top‑K retrieved candidates.


\section{Results}
\label{sec:results}

This section evaluates the proposed dataset construction and long‑term visual localization framework across the five benthic reference sites. We first validate the multi‑visit geometric registration and the footprint‑based ground truth used to relate camera views. We then characterize the footprint‑based linking of camera views, and analyze how its parameters affect the resulting link statistics. Next, we assess the performance of state‑of‑the‑art \gls{VPR} models across habitat types and revisit intervals, including within‑site spatial patterns. Finally, we compare footprint‑based and location‑based ground‑truthing strategies and examine how they influence the reported \gls{VPR} performance metrics.


\subsection{Geometric Registration}
\label{sec:results_geometric_registration}

\begin{figure}[ht]
    \centering
    \includegraphics[width=\textwidth]{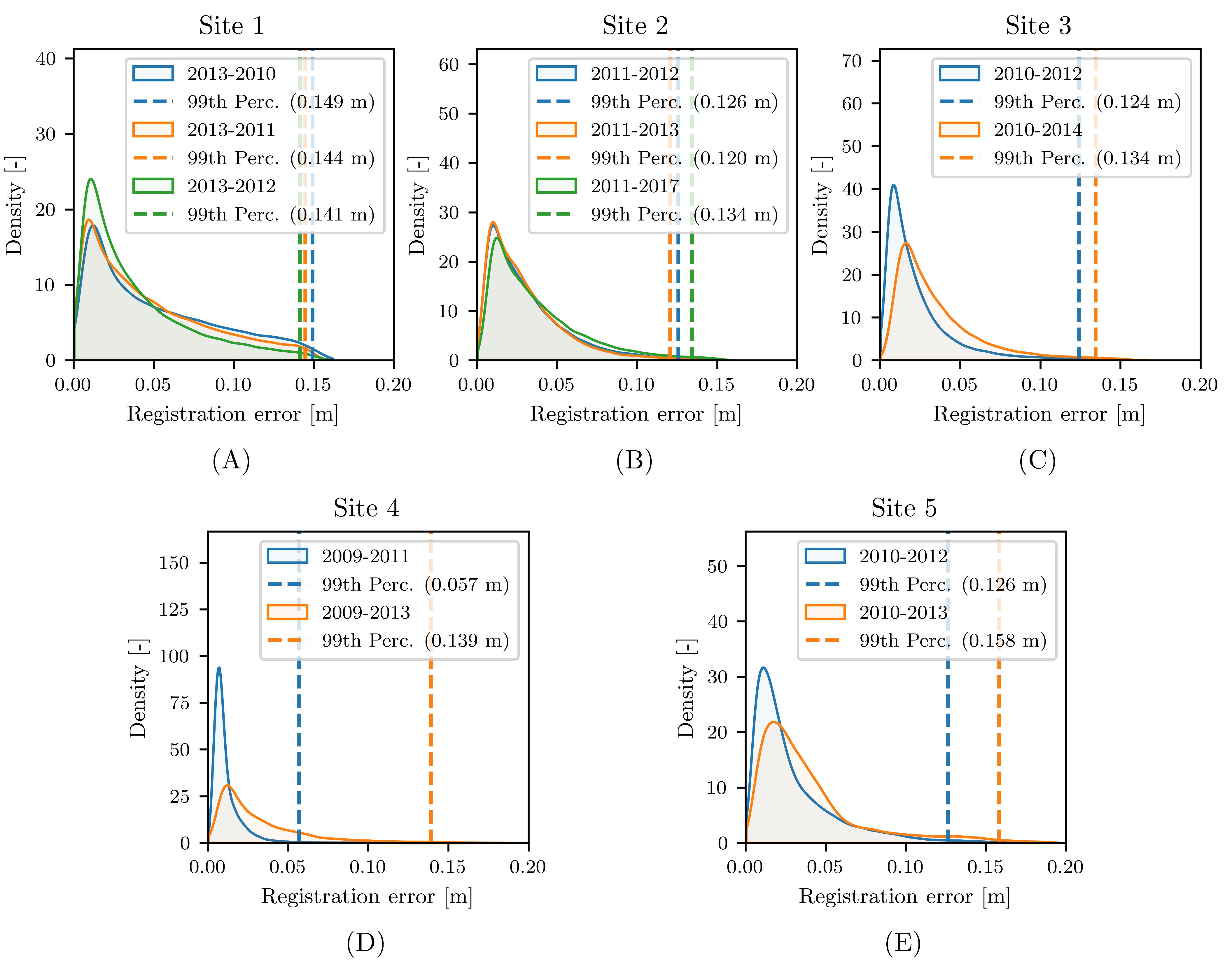}
    \caption{
        Distribution of geometric registration errors across the benthic reference sites in the dataset. The legends are on the format "Target-Source", and indicate the year of the target and source visit for each visit pair used in the geometric registration.
    }
    \label{fig:result_geometric_registration_errors}
\end{figure}

In this section, we report the geometric registration errors for the benthic reference sites, computed according to \Cref{eq:geometric_registration_distance}, for the point clouds and point correspondences from with the reconstruction and registration pipeline described in \Cref{sec:method_geometric_processing}. \Cref{fig:result_geometric_registration_errors} shows the distribution of registration errors for all target–source visit pairs at each site. Across sites, the bulk of the errors are on the order of a few centimeters, and the 99th percentile error remains below 0.16~m for all visit pairs, indicating that the visits are consistently aligned at sub-decimeter accuracy. The error distributions differ between sites: Site~1 and Site~2 exhibit heavier tails towards larger errors, whereas Sites~3–5 show more concentrated distributions with fewer large-error outliers.

\subsection{Linking Camera Views With Overlapping Image Footprints}
\label{sec:results_linking_camera_views_with_overlapping_footprints}

To estimate a conservative lower footprint~\gls{IOU} threshold, \(\footprintIouThreshold\), for which we are confident that two camera views overlap despite registration errors, we use~\Cref{eq:lower_footprint_iou_estimate}. We obtain the camera field of view along the y-axis from the calibration procedure in \Cref{sec:method_geometric_processing}, giving \(\fieldOfView_{y}=34\) deg. For the camera altitude, we set \(a=2.0\) meters, which is approximately the lowest average altitude across visits in our dataset. For the registration error, we set \(t_e=0.16\)~meters, which exceeds the 99th‑percentile registration error for all visits. With these parameter choices, \Cref{eq:lower_footprint_iou_estimate} yields a lower footprint \gls{IOU} threshold of \(\footprintIouThreshold \approx 0.07\), which we adopt for the remainder of the paper.

\Cref{tab:site_visit_pair_geometric_statistics} shows the visit pairs along with various geometry statistics across the benthic reference sites. The visit pairs have been created using the temporal pairing strategy outlined in \Cref{sec:method_evaluation_protocol_and_metrics}, and the footprint-based links have been created with the methodology in \Cref{sec:method_image_footprint_estimation}. The footprint‑based links between query and database camera views have been filtered with the lower footprint \gls{IOU} threshold ($\footprintIouThreshold = 0.07$). The query coverage overlap, defined as the fraction of query coverage area overlapped by the database visit (area intersection divided by query coverage area), ranges from approximately 50\% to 95\%. The average camera altitude varies between approximately 2.0 and 3.2~m across visit pairs and sites. The visits to Site 4 have the largest average altitude difference (\textasciitilde0.8~m between the 2009 and 2011 visits) and the highest absolute average altitude (\textasciitilde3.2~m for the 2013 visit). The average number of footprint‑based links per query camera view varies substantially across visit pairs, from about 10 to 35 links per query, reflecting differences between sites and visit pairs in terms of terrain ruggedness, survey layout, and camera altitudes. Consequently, the 95th percentile location distance between the linked camera views varies, with values ranging between 1.3 and 2.5~meters.

\setlength{\tabcolsep}{3pt} 

\begin{table}[ht]
    \centering
    \footnotesize
    \begin{tabular}{cccccccccccc}
    \toprule
    Site 
    & \makecell{Visit Years} 
    & \makecell{Database\\View Count}
    & \makecell{Query\\View Count}
    & \makecell{Avg. Database\\View Alt.}
    & \makecell{Avg. Query\\View Alt.}
    & \makecell{Query Coverage\\Overlap}
    & \makecell{Link\\Count}
    & \makecell{ALQ\textsuperscript{*}}
    & \makecell{Avg. Footprint\\Width\textsuperscript{}}
    & \makecell{Loc. Dist.\\95th Perc.\textsuperscript{}} 
    \\
    \midrule
        - & - & - & - & m & m & - & - & - & m & m \\
    \midrule
    \multirow{6}{*}{Site 1}
        & 2010 -- 2011 & 2323 & 2255 & 2.04 & 2.04 
        & 74 \% & 27.8 K & 15 & 1.65 & 1.37 \\
        & 2010 -- 2012 & 2323 & 2184 & 2.04 & 2.00 
        & 53 \% & 14.8 K & 13 & 1.65 & 1.37 \\
        & 2010 -- 2013 & 2323 & 2134 & 2.04 & 1.99 
        & 65 \% & 21.1 K & 15 & 1.64 & 1.37 \\
        & 2011 -- 2012 & 2255 & 2184 & 2.04 & 2.00 
        & 65 \% & 16.4 K & 10 & 1.64 & 1.41 \\
        & 2011 -- 2013 & 2255 & 2134 & 2.04 & 1.99 
        & 80 \% & 21.0 K & 11 & 1.63 & 1.36 \\
        & 2012 -- 2013 & 2184 & 2134 & 2.00 & 1.99 
        & 75 \% & 18.1 K & 10 & 1.62 & 1.33 \\
    \midrule
    \multirow{6}{*}{Site 2}
        & 2011 -- 2012 & 2317 & 2225 & 2.27 & 2.10 
        & 80 \% & 23.9 K & 14 & 1.82 & 1.99 \\
        & 2011 -- 2013 & 2317 & 2164 & 2.27 & 2.00 
        & 87 \% & 24.7 K & 13 & 1.79 & 2.04 \\
        & 2011 -- 2017 & 2317 & 3467 & 2.27 & 2.42 
        & 82 \% & 50.4 K & 16 & 1.95 & 2.32 \\
        & 2012 -- 2013 & 2225 & 2164 & 2.10 & 2.00 
        & 82 \% & 23.9 K & 11 & 1.69 & 1.65 \\
        & 2012 -- 2017 & 2225 & 3467 & 2.10 & 2.42 
        & 72 \% & 34.1 K & 12 & 1.86 & 2.18 \\
        & 2013 -- 2017 & 2164 & 3467 & 2.00 & 2.42 
        & 75 \% & 31.4 K & 11 & 1.83 & 2.25 \\
    \midrule
    \multirow{3}{*}{Site 3}
        & 2010 -- 2012 & 2728 & 2597 & 2.14 & 2.01 
        & 66 \% & 24.4 K & 12 & 1.64 & 1.47 \\
        & 2010 -- 2014 & 2728 & 2246 & 2.14 & 2.24 
        & 61 \% & 20.0 K & 13 & 1.71 & 1.54 \\
        & 2012 -- 2014 & 2597 & 2246 & 2.01 & 2.24 
        & 66 \% & 18.7 K & 11 & 1.67 & 1.56 \\
    \midrule
    \multirow{3}{*}{Site 4}
        & 2009 -- 2011 & 6280 & 2267 & 3.24 & 2.48 
        & 96 \% & 77.8 K & 35 & 2.26 & 2.48 \\
        & 2009 -- 2013 & 6280 & 2134 & 3.24 & 2.26 
        & 79 \% & 60.8 K & 31 & 2.07 & 2.46 \\
        & 2011 -- 2013 & 2267 & 2134 & 2.48 & 2.26 
        & 57 \% & 19.4 K & 15 & 1.82 & 1.97 \\
    \midrule
    \multirow{3}{*}{Site 5}
        & 2010 -- 2012 & 2701 & 3687 & 2.20 & 2.13 
        & 76 \% & 50.5 K & 16 & 1.82 & 1.87 \\
        & 2010 -- 2013 & 2701 & 3313 & 2.20 & 2.32 
        & 82 \% & 49.0 K & 17 & 1.87 & 1.85 \\
        & 2012 -- 2013 & 3687 & 3313 & 2.13 & 2.32 
        & 87 \% & 73.2 K & 24 & 1.88 & 1.95 \\
    \bottomrule
    \end{tabular}
    \caption{
        Overview of the visit pairs and footprint links filtered with $\footprintIouThreshold = 0.07$. The visit pairs are created using the temporal pairing strategy outlined in~\Cref{sec:method_evaluation_protocol_and_metrics}. \textsuperscript{*}Average link per valid query.
    }
    \label{tab:site_visit_pair_geometric_statistics}
\end{table}


\subsection{Evaluating Long-Term Visual Place Recognition Performance}
\label{sec:results_evaluating_long_term_vpr_performance}

In this subsection, we evaluate long-term \gls{VPR} performance across all reference sites and visit pairs in our dataset using our footprint-based ground truth. We first compare how \gls{SOTA} \gls{VPR} models perform across different seafloor types  (\Cref{sec:results_evaluating_vpr_performance_across_sites}) and analyze within-site spatial patterns for the localization results of a selected model (\Cref{sec:results_within_site_spatial_vpr_patterns}). We then analyze how model performance varies with different revisit intervals for each site (\Cref{sec:results_vpr_performance_across_revisits}).


\subsubsection{Evaluating Visual Place Recognition Performance Across Sites}
\label{sec:results_evaluating_vpr_performance_across_sites}

\begin{figure}[ht]
    \centering
    \includegraphics[width=\textwidth]{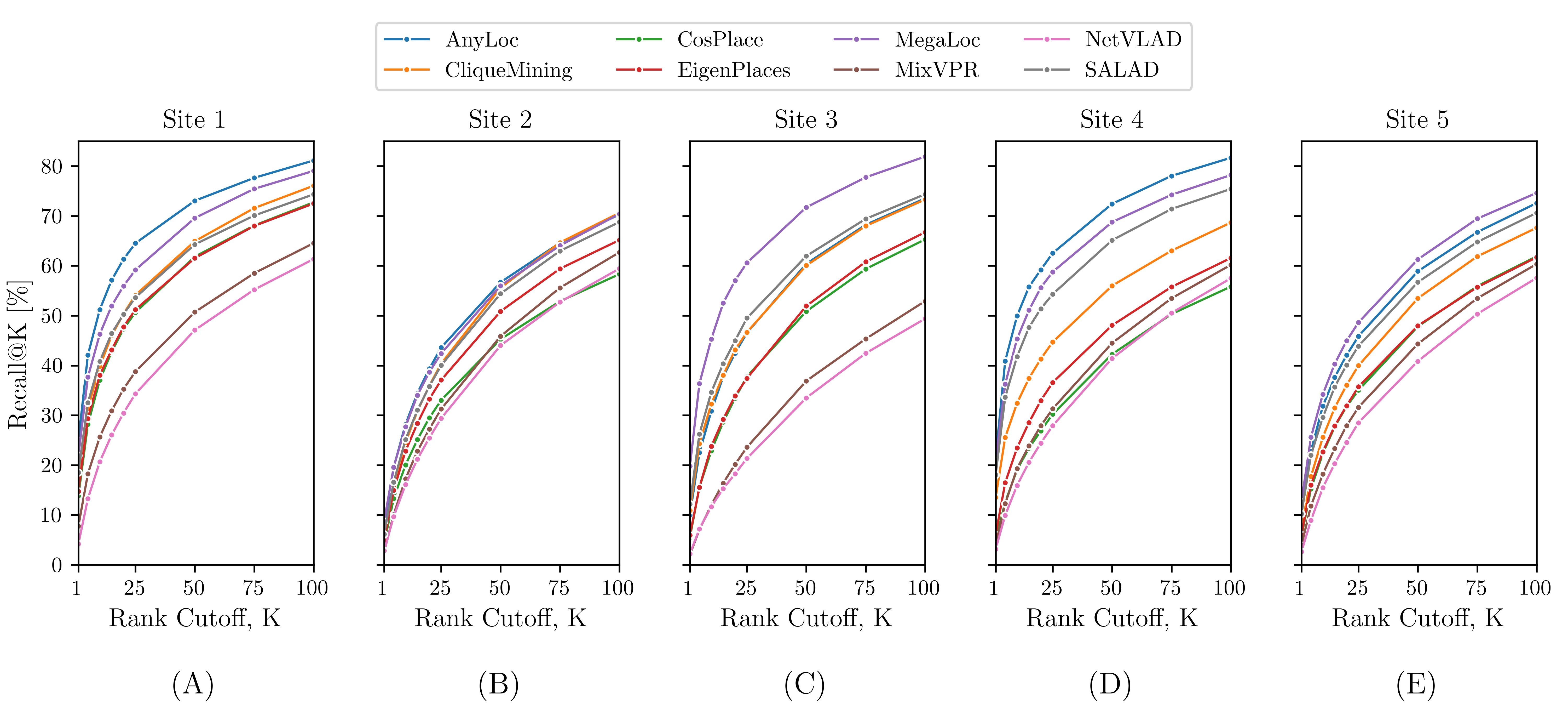}
    \caption{
        Mean Recall@K across all visit pairs for each site in the dataset. Results are based on the footprint-based ground truth from \Cref{sec:results_linking_camera_views_with_overlapping_footprints}.
    }
    \label{fig:mean_site_recall_curves}
\end{figure}

To assess how the selected \gls{VPR} models generalize across seafloor types, we evaluate their mean $\metricVprRecallK$ across all visit pairs for each reference site. This provides insight into how seafloor characteristics influence localization performance. As shown in \Cref{fig:mean_site_recall_curves}, we report the mean $\metricVprRecallK$ over a range of rank cutoffs $\rankCutoff$, illustrating how retrieval accuracy varies with the number of candidates considered. Ground-truth correspondences are defined using footprint-based camera links established through the methodology in \Cref{sec:method_evaluation_protocol_and_metrics}, applying the lower footprint \gls{IOU} threshold ($\footprintIouThreshold=0.07$) derived in \Cref{sec:results_linking_camera_views_with_overlapping_footprints} to compensate for residual registration errors between visits.

The mean recall curves show significant variation in \gls{VPR} performance across sites. Site 1, Site 3, and Site 4 exhibit higher mean recall across the entire range of $\rankCutoff$ compared to the other sites. Recall is moderately lower at Site 5, although the separation between models remains clear, whereas Site 2 yields the lowest recall with only minor differences between models. The relative differences in recall between higher- and lower-performing models are most clear at lower rank cutoffs $\rankCutoff$. AnyLoc and MegaLoc are generally the two strongest models and consistently achieve the highest recall across most sites. MegaLoc is particularly strong at Site 3, where it achieves considerably higher recall than all other models. We also observe that the \gls{VIT}-based models (AnyLoc, CliqueMining, MegaLoc, and SALAD) consistently achieve higher recall than the \gls{CNN}-based models (CosPlace, EigenPlaces, MixVPR, and NetVLAD) across all sites.

For practical \gls{VPR} applications, performance at lower rank cutoffs is particularly relevant, since the system typically only considers a small set of top-ranked candidates for downstream verification. Consequently, we focus on two representative operating points and report the mean Recall@1 and Recall@10 over all visit pairs for each site in \Cref{tab:mean_place_recognition_recall}. This complements the trends observed in \Cref{fig:mean_site_recall_curves} by quantifying how often each model retrieves the correct place as the top match or within the top 10 candidates. 

Overall, the numbers in \Cref{tab:mean_place_recognition_recall} confirm the trends observed in \Cref{fig:mean_site_recall_curves}, with AnyLoc and MegaLoc achieving the highest Recall@1 and Recall@10 on most sites. In particular, AnyLoc performs slightly better than MegaLoc on Site 1, achieving Recall@1 and Recall@10 of 24.9\% and 51.2\%, respectively, compared to 21.9\% and 46.3\% for MegaLoc. Conversely, MegaLoc is considerably stronger on Site 3, with Recall@1 and Recall@10 of 19.8\% and 45.3\%, whereas AnyLoc achieves 10.0\% and 30.8\%, respectively. A table with Recall@1 and Recall@10 for all visit pairs in the dataset can be found in Supplementary Materials.

\newcommand{\bestRecall}[1]{\textcolor{green!60!black}{#1}}
\newcommand{\nextBestRecall}[1]{\textcolor{red!70!black}{#1}}

\begin{table*}[t]
    \centering
    \footnotesize
    \begin{tabular}{l cc cc cc cc cc}
    \toprule
    \multirow[c]{2}{*}{Methods}
     & \multicolumn{2}{c}{Site 1} 
     & \multicolumn{2}{c}{Site 2} 
     & \multicolumn{2}{c}{Site 3}
     & \multicolumn{2}{c}{Site 4} 
     & \multicolumn{2}{c}{Site 5} \\
    \cmidrule(l{3pt}r{3pt}){2-3}
    \cmidrule(l{3pt}r{3pt}){4-5}
    \cmidrule(l{3pt}r{3pt}){6-7}
    \cmidrule(l{3pt}r{3pt}){8-9}
    \cmidrule(l{3pt}r{3pt}){10-11}
     & R@1     & R@10
     & R@1     & R@10
     & R@1     & R@10
     & R@1     & R@10
     & R@1     & R@10 \\
    \midrule
    & \% & \% & \% & \% & \% & \% & \% & \% & \% & \% \\
    \midrule
    AnyLoc \citep{keetha_anyloc_2024}
        & \bestRecall{24.9} & \bestRecall{51.2} 
        & \nextBestRecall{7.4} & \bestRecall{28.2} 
        & 10.0 & 30.8 
        & \bestRecall{21.4} & \bestRecall{49.9} 
        & 10.1 & \nextBestRecall{31.9} \\
    CliqueMining \citep{izquierdo_close_2025}
        & 17.0 & 39.8
        & 5.6 & 24.6
        & 10.9 & 32.3 
        & 13.5 & 32.4 
        & 7.2 & 25.6 \\
    CosPlace \citep{zaccone_distributed_2024}
        & 13.8 & 37.1 
        & 4.3 & 20.1
        & 5.5 & 22.9 
        & 4.9 & 19.1 
        & 5.7 & 22.3 \\
    EigenPlaces \citep{berton_eigenplaces_2023}
        & 14.7 & 38.0
        & 4.9 & 22.8 
        & 6.0 & 22.7
        & 6.0 & 23.5 
        & 5.9 & 23.8 \\
    MegaLoc \citep{berton_megaloc_2025}
        & \nextBestRecall{21.9} & \nextBestRecall{46.3} 
        & \bestRecall{8.7} & \nextBestRecall{27.7} 
        & \bestRecall{19.8} & \bestRecall{45.3} 
        & \nextBestRecall{20.9} & \nextBestRecall{45.3} 
        & \bestRecall{12.6} & \bestRecall{34.2} \\
    MixVPR \citep{ali-bey_mixvpr_2023}
        & 7.8 & 25.6 
        & 2.6 & 17.3 
        & 2.0 & 12.0 
        & 4.0 & 19.3 
        & 3.9 & 18.2 \\
    NetVLAD \citep{arandjelovic_netvlad_2016}
        & 4.2 & 20.7
        & 2.8 & 16.1 
        & 2.2 & 11.7 
        & 3.1 & 15.9 
        & 2.6 & 15.5 \\
    SALAD \citep{izquierdo_optimal_2024}
        & 18.5 & 40.8
        & 6.1 & 25.1
        & \nextBestRecall{12.2} & \nextBestRecall{34.7}
        & 18.1 & 41.8 
        & \nextBestRecall{10.2} & 29.6 \\
    \bottomrule
    \end{tabular}
    \caption{
        Recall@1 and Recall@10 averaged across every visit pair for the reference sites in the dataset. For each rank cutoff $K$ and site, we use \bestRecall{green} and \nextBestRecall{red} to highlight the best and second best model, respectively. Results are based on the footprint-based ground truth from \Cref{sec:results_linking_camera_views_with_overlapping_footprints}.
    }
    \label{tab:mean_place_recognition_recall}
\end{table*}


\subsubsection{Within-Site Spatial Visual Place Recognition Patterns}
\label{sec:results_within_site_spatial_vpr_patterns}

In order to highlight spatial patterns in \gls{VPR} performance attributable to seafloor and motion characteristics, we present maps with the localization results for MegaLoc, which is the best overall performing model according to \Cref{sec:results_evaluating_vpr_performance_across_sites}. We show localization results using a limited number of \gls{VPR} candidates per query image, with a rank cutoff of $\rankCutoff=5$. This reflects a setting where only a small number of top-ranked candidates are considered for downstream verification.

\begin{figure}[!ht]
    \centering
    \includegraphics[width=\textwidth]{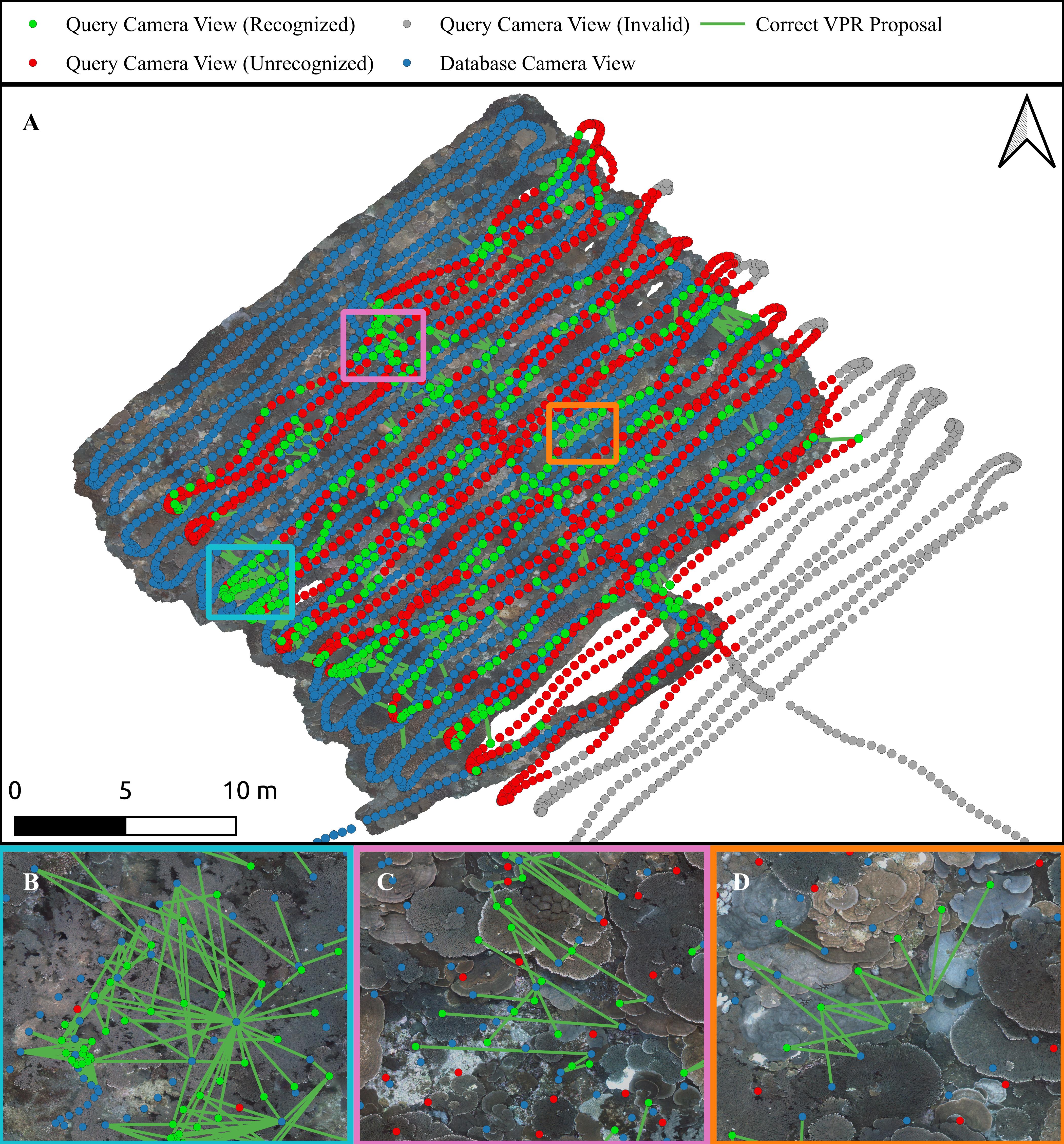}
    \caption{
        Visual place recognition results at Site 1 rendered on top of an orthomosaic derived from the 2010 images, using MegaLoc with $\rankCutoff=5$ to retrieve 2010 database images for 2013 query images. (A) Overview of place recognition results for the entire site. (B) Cluster of recognized query camera views at the end of a survey leg. (C) Cluster of recognized query camera views at a cross-over point between survey legs. (D) Trajectory segment with multiple consecutively recognized query camera views.
    }
    \label{fig:localization_map_site1_2010_2013}
\end{figure}

\begin{figure}[!ht]
    \centering
    \includegraphics[width=\textwidth]{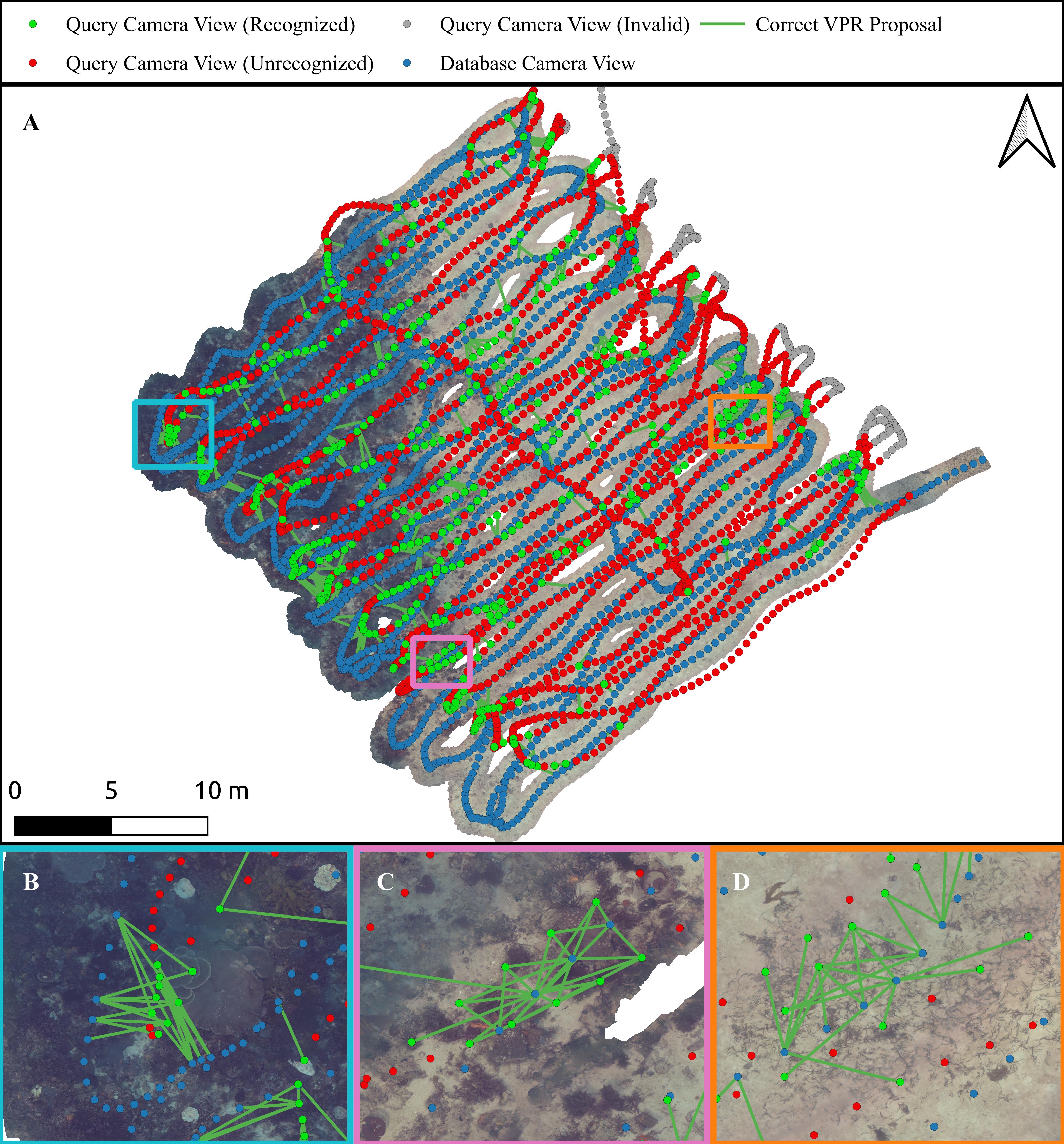}
    \caption{
        Visual place recognition results at Site 2 rendered on top of an orthomosaic derived from the 2011 images, using MegaLoc with $\rankCutoff=5$ to retrieve 2011 database images for 2013 query images. Panel (A) show an overview of place recognition results for the entire site. Panel (B) shows a cluster of recognized query camera views at the end of a survey leg. Panel (C) shows trajectory segments with recognized query camera views as the \gls{AUV} passes over coral colonies in the transition between the soft sediment bottom and the dense coral reef. Panel (D) shows trajectory segments with recognized query camera views as the \gls{AUV} passes over a seagrass meadow on the soft sediment bottom.
    }
    \label{fig:localization_map_site2_2011_2013}
\end{figure}

\Cref{fig:localization_map_site1_2010_2013} shows a map with the localization results for the 2010–2013 visit pair to Site 1 using MegaLoc. Each query camera view is labeled as recognized (at least one correct proposal in the top $\rankCutoff$), unrecognized (no correct proposals), or invalid (no ground-truth match), according to the protocol in \Cref{sec:method_evaluation_protocol_and_metrics}. Correct \gls{VPR} proposals are visualized as links between recognized query views and their matched database views. \Cref{fig:localization_map_site2_2011_2013} shows the corresponding localization results for the 2011–2013 visit pair to Site 2.

In \Cref{fig:localization_map_site1_2010_2013}(A), recognized queries are distributed throughout Site 1 but typically appear in spatial clusters of 2–5 consecutive views along the trajectory. Panel (B) shows a zoom-in on the end of a survey leg where the vehicle turns, with a high concentration of recognized queries. In (C), recognized queries cluster at a cross-over point between survey legs. In (D), multiple consecutive recognized queries are matched to a parallel trajectory segment.

\Cref{fig:localization_map_site2_2011_2013}(A) provides an overview of the localization results throughout Site 2. The western part of the site covers a dense coral reef, the eastern part consists of soft sediment bottom, and a transition zone with sparse corals lies between them. The overview reveals a clear difference in localization performance between these regions, where recognized queries are much more frequent over areas with corals, whereas they are less common over the soft sediment bottom. Panel (B) shows large concentrations of recognized queries at a survey leg end, similar to Site 1. In (C), multiple trajectory segments with recognized queries appear over seafloor patches with sparse corals in the transition zone. In (D), trajectory segments with recognized queries lie over a seagrass meadow on the soft sediment bottom.

Across both \Cref{fig:localization_map_site1_2010_2013} and \Cref{fig:localization_map_site2_2011_2013}, recognized queries are typically spatially clustered, either around patches with distinctive seafloor features or along trajectory segments where the motion produces many images with overlapping footprints. Similar maps with localization results for visit pairs to Site 3, 4, and 5 can be found in Supplementary Materials.

\subsubsection{Evaluating Visual Place Recognition Performance Across Revisits}
\label{sec:results_vpr_performance_across_revisits}

\begin{figure}[ht]
    \centering
    \includegraphics[width=\textwidth]{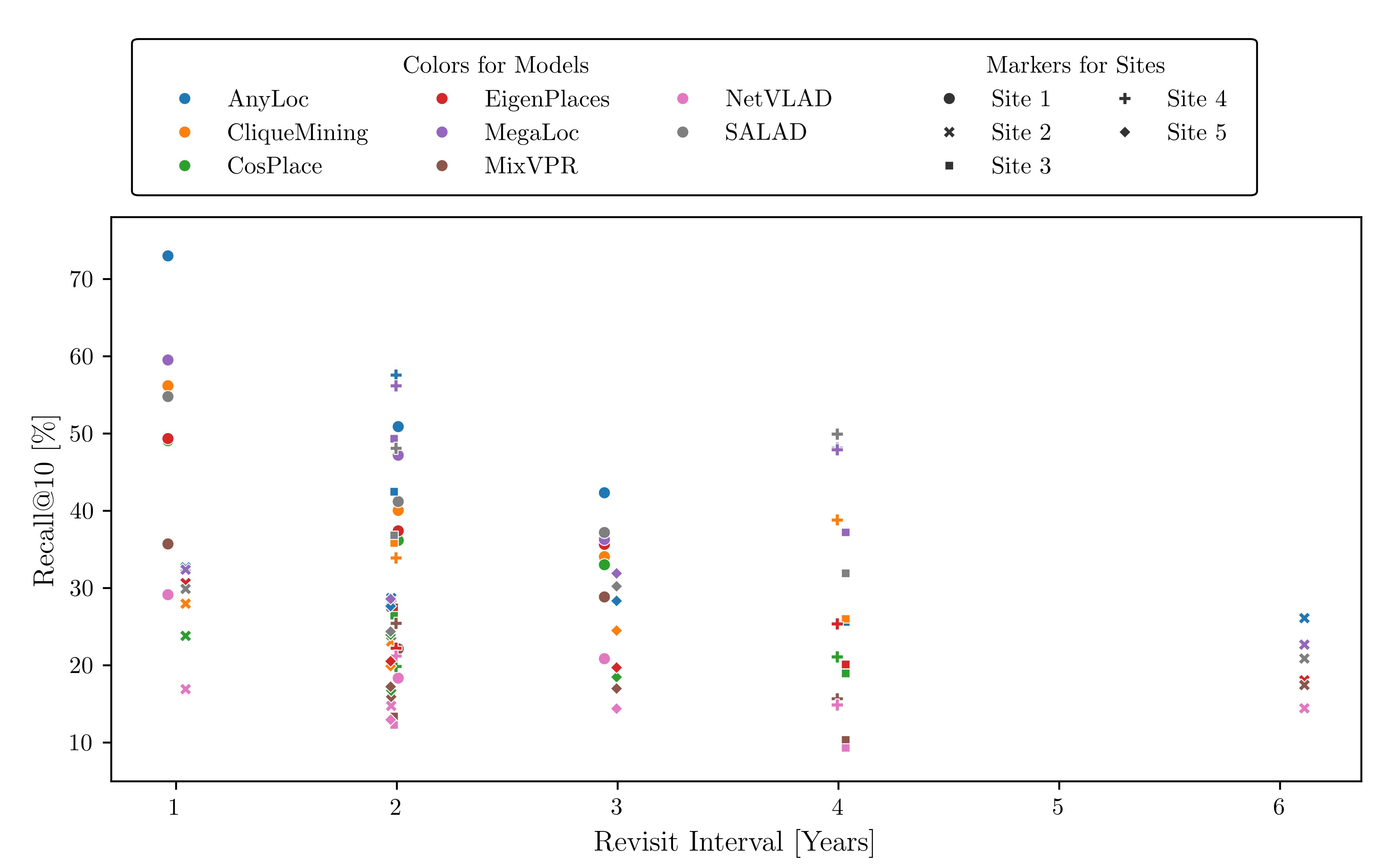}
    \caption{
        Recall@10 for all \gls{VPR} models for varying revisit intervals to the sites in the dataset. The overall trend shows a decline of \gls{VPR} Recall@10 with increasing revisit intervals. For Site 1 and Site 2, Recall@10 shows a sharp decline between revisit intervals of 1 and 2 years, before a more gradual decline. For Site 5, Recall@10 increases between revisit intervals of 2 and 3 years, indicating that additional factors affect Recall@10. Results are based on the footprint-based ground truth from \Cref{sec:results_linking_camera_views_with_overlapping_footprints}.
    }
    \label{fig:recall@10_vs_revisit_interval_scatter_plot}
\end{figure}

To evaluate how \gls{VPR} performance is affected by increasing revisit interval, i.e., the time between two visits, we report Recall@10 for visit pairs where the first visit to each site serves as the database visit. In \Cref{fig:recall@10_vs_revisit_interval_scatter_plot}, the general trend is that Recall@10 decreases with increasing revisit interval. For the sites with lower overall recall, Site 2 and Site 5, this trend is less pronounced, and Recall@10 remains at similar levels across a range of revisit intervals. For sites with more than three visits, i.e., Site 1 and Site 2, we observe a sharper decline between revisit intervals of 1 and 2 years, followed by a more gradual decline towards a plateau at longer intervals.

\subsection{Evaluating Differences in Ground-Truthing Strategies}
\label{sec:results_evaluating_differences_in_ground_truthing_strategies}

To quantify how footprint-based and location-based ground-truthing strategies affect \gls{VPR} performance metrics, we analyze the performance of the \gls{VPR} models under two ground-truth definitions, one footprint-based and one location-based. For the footprint-based ground truth, we use the lower footprint \gls{IOU} threshold ($\footprintIouThreshold = 0.07$) as established in \Cref{sec:results_linking_camera_views_with_overlapping_footprints}. For each visit pair, we set the distance threshold for the location-based ground truth to the 95th percentile of distances between footprint-linked camera views. This approach makes the location-based ground truth for each visit pair largely overlap with the footprint-based one, while excluding the most extreme spatial outliers. However, because the location-based ground truth does not explicitly enforce footprint overlap or camera field of view constraints, it also introduces additional links that the footprint-based method deliberately excludes (see Panel (A) in \Cref{fig:method_footprint_estimation_illustration}).

To compare the \gls{VPR} models for the two ground truths, we use the \gls{VPR}-specific definition of recall, $\metricVprRecallK$, defined in~\Cref{eq:vpr_recall@K}, and additionally the information‑retrieval definition of recall, $\metricInfoRecallK$, defined in~\Cref{eq:information_recall@K}. We use both recall definitions because \gls{VPR} recall is not penalized for false negatives since adding more ground-truth links can only increase the probability that a query is counted as recognized. In contrast, information‑retrieval recall $\metricInfoRecallK$ is penalized for false negatives, and therefore decreases when additional relevant query–database pairs are not retrieved. The \gls{VPR} recall $\metricVprRecallK$ and information-retrieval recall $\metricInfoRecallK$ for $\rankCutoff=10$ are denoted R@10 and IR@10, respectively.

\begin{figure}
    \centering
    \includegraphics[width=\linewidth]{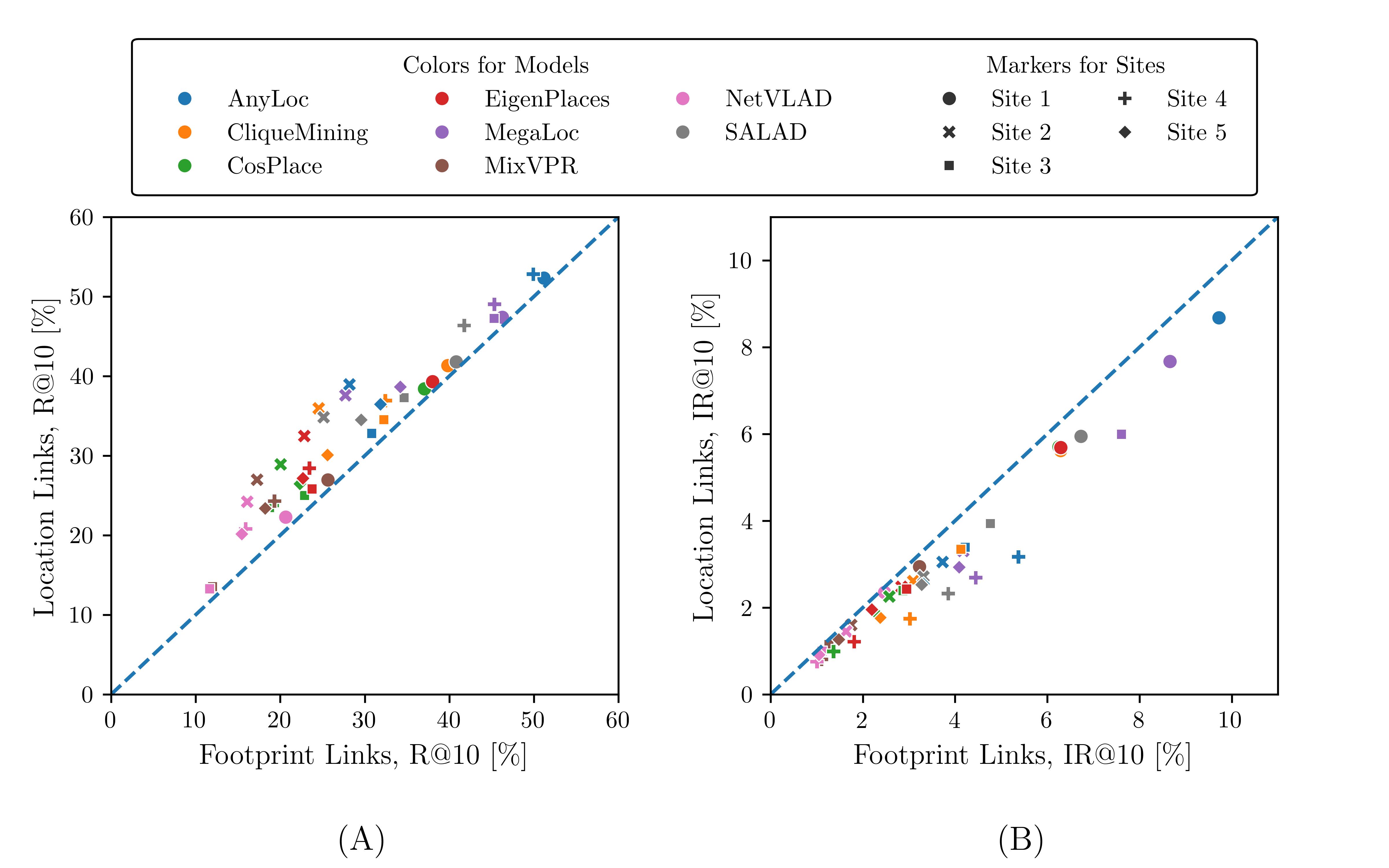}
    \caption{
        Scatter plots of (A) R@10 and (B) IR@10 for the footprint-based and location-based ground truth. The R@10 is consistently higher for the location-based ground truth for all sites and \gls{VPR} models, and indicates that model performance is over-estimated with the location-based ground truth. IR@10 is consistently higher for the footprint-based ground truth for all sites and models, and indicate that the \gls{VPR} retrievals are more aligned with the footprint-based links.
    }
    \label{fig:results_ground_truth_comparison_scatter}
\end{figure}

\Cref{fig:results_ground_truth_comparison_scatter} shows R@10 (A) and IR@10 (B) across all \gls{VPR} models for the footprint-based and location-based ground truths, averaged over site visit pairs. R@10 is consistently higher under the location-based ground truth, whereas IR@10 is consistently higher under the footprint-based ground truth, reflecting the different ways the two recall definitions respond to additional ground-truth links. Examining the R@10 scatter plot, points for each site cluster along lines with a roughly constant offset from the diagonal. This pattern indicates that the R@10 offsets between the two ground truths are largely specific to each site and relatively consistent across both strong and weak \gls{VPR} models. For IR@10, the points do not follow a roughly constant offset from the diagonal. Instead, stronger models exhibit a larger deviation from the diagonal than weaker models, indicating that they retrieve a larger portion of the footprint-based links than of the additional links introduced by the location-based ground truth.

To relate the differences in performance metrics to the geometry of the two ground truths, we report the metrics for MegaLoc and NetVLAD, which are considered the strongest and weakest model, for each visit pair together with key geometric statistics for the footprint-based and location-based ground truth. \Cref{tab:results_ground_truth_comparison} shows the average footprint widths and distance thresholds, and the average number of links per valid query (\gls{ALQ}) for the two ground truths. Across sites, the location-based ground truth yields a higher \gls{ALQ} than the footprint-based ground truth. At Site 1, the distance threshold is close to the effective footprint width and \gls{ALQ} increases only slightly, whereas at Sites 2, 4, and 5 the threshold substantially exceeds the footprint width and location-based \gls{ALQ} increases markedly. The overall higher \gls{ALQ} for the location-based ground truth indicates that it includes numerous query–database links that are not present under the more conservative footprint-based ground truth.

For R@10, both models achieve consistently higher scores under the location-based than under the footprint-based ground truth, whereas IR@10 is generally lower under the location-based ground truth, consistent with the patterns observed in \Cref{fig:results_ground_truth_comparison_scatter}. \Cref{tab:results_ground_truth_comparison} further shows that visit pairs with the largest increase in R@10 when switching from footprint-based to location-based ground truth typically correspond to cases where the distance threshold is substantially larger than the average footprint width. This pattern is most evident at Site 2. For the 2012–2013 visit pair at Site 2, MegaLoc’s R@10 increases only modestly from 29.0\% to 33.4\% when moving from footprint-based to location-based ground truth, and the distance threshold (1.65~m) is very close to the average footprint width (1.69~m). In contrast, for the 2013–2017 visit pair, the distance threshold (2.25~m) exceeds the average footprint width (1.83~m) by a larger margin, and MegaLoc’s R@10 increases more substantially, from 30.8\% to 41.8\%. A similar pattern is observed for NetVLAD, with lower absolute R@10 values. In line with the IR@10 patterns, the additional location-based links introduced when the distance threshold substantially exceeds the footprint width tend to reduce IR@10, particularly for MegaLoc.

\begin{table}[!ht]
    \centering
    \footnotesize
    \begin{tabular}{cccccccccccccccc}
    \toprule
    \multirow[c]{2}{*}{\centering Site}
    & \multirow[c]{2}{*}{\centering \makecell{Visit Years}}
    & \multirow[c]{2}{*}{\centering \makecell{Foot.\\Width}}
    & \multirow[c]{2}{*}{\centering \makecell{Dist.\\Thres.}}
    & \multirow[c]{2}{*}{\centering \makecell{Foot.\\ALQ}}
    & \multirow[c]{2}{*}{\centering \makecell{Loc.\\ALQ}}
    & \multicolumn{2}{c}{\makecell{MegaLoc\\Foot. Links}}
    & \multicolumn{2}{c}{\makecell{MegaLoc\\Loc. Links}}
    & \multicolumn{2}{c}{\makecell{NetVLAD\\Foot. Links}}
    & \multicolumn{2}{c}{\makecell{NetVLAD\\Loc. Links}}
    \\
    \cmidrule(l{2pt}r{2pt}){7-8}
    \cmidrule(l{2pt}r{2pt}){9-10}
    \cmidrule(l{2pt}r{2pt}){11-12}
    \cmidrule(l{2pt}r{2pt}){13-14}
    & 
    &
    &
    & 
    & 
    & IR@10
    & R@10
    & IR@10
    & R@10
    & IR@10
    & R@10
    & IR@10
    & R@10
    \\
    \midrule
     - & - & m & m & - & - & \% & \% & \% & \% & \% & \% & \% & \% \\
    \midrule
    \multirow{6}{*}{Site 1}
        & 2010 -- 2011 & 1.65 & 1.37 & 15 & 18 
        & 9.8 & 59.5 & 8.7 & 60.5 & 3.1 & 29.1 & 2.9 & 31.2 \\ 
        & 2010 -- 2012 & 1.65 & 1.37 & 13 & 15 
        & 6.8 & 47.2 & 6.1 & 49.1 & 1.7 & 18.3 & 1.6 & 19.4 \\ 
        & 2010 -- 2013 & 1.64 & 1.37 & 15 & 17 
        & 5.0 & 36.3 & 4.5 & 37.8 & 1.8 & 20.9 & 1.8 & 22.7 \\ 
        & 2011 -- 2012 & 1.64 & 1.41 & 10 & 13
        & 9.6 & 46.1 & 8.1 & 46.7 & 2.6 & 16.8 & 2.6 & 19.2 \\ 
        & 2011 -- 2013 & 1.63 & 1.36 & 11 & 13 
        & 6.7 & 33.9 & 6.1 & 35.2 & 2.9 & 18.5 & 2.7 & 19.6 \\ 
        & 2012 -- 2013 & 1.62 & 1.33 & 10 & 12 
        & 14.1 & 54.7 & 12.5 & 55.0 & 2.7 & 20.3 & 2.5 & 21.6 \\ 
    \midrule
    \multirow{6}{*}{Site 2}
        & 2011 -- 2012 & 1.82 & 1.99 & 14 & 26 
        & 4.5 & 32.4 & 3.4 & 42.0 & 1.6 & 16.9 & 1.5 & 26.1 \\
        & 2011 -- 2013 & 1.79 & 2.04 & 13 & 27 
        & 3.8 & 27.6 & 2.9 & 40.0 & 1.4 & 14.7 & 1.2 & 23.0 \\
        & 2011 -- 2017 & 1.95 & 2.32 & 16 & 35 
        & 2.4 & 22.7 & 2.1 & 36.4 & 1.2 & 14.4 & 1.1 & 21.7 \\
        & 2012 -- 2013 & 1.69 & 1.65 & 11 & 18 
        & 5.5 & 29.0 & 4.8 & 33.4 & 1.9 & 16.7 & 1.7 & 21.1 \\
        & 2012 -- 2017 & 1.86 & 2.18 & 12 & 27 
        & 3.5 & 23.9 & 2.7 & 32.0 & 1.5 & 13.9 & 1.4 & 22.5 \\
        & 2013 -- 2017 & 1.83 & 2.25 & 11 & 28 
        & 5.5 & 30.8 & 3.9 & 41.8 & 2.3 & 20.0 & 1.9 & 30.9 \\
    \midrule
    \multirow{3}{*}{Site 3}
        & 2010 -- 2012 & 1.64 & 1.47 & 12 & 16
        & 8.6 & 49.3 & 6.6 & 51.3 & 1.2 & 12.3 & 1.0 & 13.7 \\
        & 2010 -- 2014 & 1.71 & 1.54 & 13 & 17 
        & 5.5 & 37.2 & 4.6 & 39.4 & 0.7 & 9.3 & 0.7 & 10.9 \\
        & 2012 -- 2014 & 1.67 & 1.56 & 11 & 15 
        & 8.7 & 49.4 & 6.8 & 51.1 & 1.4 & 13.4 & 1.3 & 15.3 \\
    \midrule
    \multirow{3}{*}{Site 4}
        & 2009 -- 2011 & 2.26 & 2.48 & 35 & 63 
        & 4.2 & 56.2 & 2.6 & 60.5 & 1.0 & 21.2 & 0.7 & 25.6 \\
        & 2009 -- 2013 & 2.07 & 2.46 & 31 & 57 
        & 4.3 & 47.9 & 2.3 & 51.4 & 0.8 & 14.9 & 0.5 & 19.7 \\
        & 2011 -- 2013 & 1.82 & 1.97 & 15 & 25 
        & 4.9 & 31.9 & 3.2 & 35.3 & 1.3 & 11.7 & 1.1 & 17.1 \\
    \midrule
    \multirow{3}{*}{Site 5}
        & 2010 -- 2012 & 1.82 & 1.87 & 16 & 27
        & 3.6 & 28.6 & 2.6 & 32.8 & 1.0 & 13.0 & 0.9 & 17.6 \\
        & 2010 -- 2013 & 1.87 & 1.85 & 17 & 26 
        & 4.0 & 31.9 & 3.2 & 37.0 & 1.1 & 14.4 & 0.9 & 18.5 \\
        & 2012 -- 2013 & 1.88 & 1.95 & 24 & 39 
        & 4.7 & 42.2 & 3.0 & 46.1 & 1.1 & 19.0 & 1.0 & 24.4 \\
    \bottomrule
    \end{tabular}
    \caption{
        Summary of geometric statistics and retrieval performance for MegaLoc and NetVLAD under footprint-based and location-based ground truth. 
        Footprint-based ground-truth links from \Cref{sec:results_linking_camera_views_with_overlapping_footprints}. For each visit pair, the location-based distance threshold is set to the 95th percentile of distances between footprint-linked camera views. “Foot. Width” and “Dist. Thres.” give the average footprint width and the resulting distance threshold, respectively. “Foot. ALQ” and “Loc. ALQ” report the average number of links per valid query under the two ground truths. R@10 and IR@10 denote $\metricVprRecallK$ and $\metricInfoRecallK$ at $\rankCutoff = 10$, respectively.
    }
    \label{tab:results_ground_truth_comparison}
\end{table}


\section{Discussion}
\label{sec:discussion}


\subsection{Geometric Registration}
\label{sec:discussion_geometric_registration}

Across all reference sites, the geometric registration between visits achieves sub-decimeter alignment, with occasional local errors up to about $0.16$~m that are not representative of the overall registration quality (see \Cref{sec:results_geometric_registration}). The largest errors arise mainly from structural habitat changes in dense coral reef areas at Sites~1 and~2, where genuine 3D differences appear as local registration errors, and from missing overlap or reconstruction holes, where the registration algorithm may associate invalid point correspondences that inflate local error metrics without reflecting alignment quality where overlap is present. We also observe mild low-frequency warping in some reconstructions, likely caused by accumulated calibration and pose-estimation errors. For the approximately $35\times35$~m reference sites considered here, this typically contributes only a few centimeters of error, but larger sites might require non-rigid registration to align visits \citep{monji-azad_review_2023, wang_strategies_2022}.

Overall, the largest registration errors can therefore be traced to local structural changes, areas of missing overlap, or mild global warping rather than to systematic misalignment between visits. Consequently, the reported sub-decimeter error range is a more representative measure of the achieved alignment quality, and this level of consistency is sufficient to treat repeated surveys as sharing a common reference frame, enabling validation of long-term visual localization results at sub-decimeter accuracy.


\subsection{Linking Camera Views With Overlapping Image Footprints}
\label{sec:discussion_linking_camera_views_with_overlapping_footprints}

Recent work defines \gls{VPR} as the ability to recognize one’s location based on reference and query observations perceived from overlapping fields of view, implying that successful matches require a certain degree of visual overlap between observations \citep{garg_where_2021}. In this view, our footprint-based ground truth directly encodes this notion of place by linking camera views only when their estimated seafloor footprints overlap.

The results in \Cref{sec:results_linking_camera_views_with_overlapping_footprints} demonstrate that the footprint estimation and overlap criterion reliably establish cross-visit links between camera views, supporting the robustness of our footprint-based linking approach under varying geometry. The lower footprint \gls{IOU} threshold used in this work ($\footprintIouThreshold = 0.07$) is a deliberately conservative choice. It is based on the largest observed registration errors, which, as discussed in \Cref{sec:discussion_geometric_registration}, are not representative of typical misalignment between visits. As a result, some camera view pairs with minimal but non-negligible visual overlap are likely excluded. However, this conservative threshold ensures that retained links exhibit meaningful visual overlap while minimizing false matches due to residual registration errors.

Across the sites and visit pairs, the relative camera–seafloor geometry is strongly affected by terrain structure and vehicle motion. Two dominant contributors are terrain ruggedness and vehicle altitude relative to the seafloor. Rugged terrain can limit the effective field of view and cause the image footprint to change substantially between neighboring views, so visual overlap becomes highly sensitive to local camera–seafloor geometry (see \Cref{fig:method_footprint_estimation_illustration}, Panel (A)). This is the case at Site 4, which covers a boulder reef where images can exhibit range variations of several meters within a single field of view (see \Cref{fig:range_map_fusion_example}). Local vertical seafloor structure can also interact with the vehicle’s altitude control and obstacle avoidance to produce distinct high- and low-altitude image sets (see \Cref{fig:method_footprint_estimation_illustration}, Panel (B)). This occurs, for example, at Site 2, where a dense coral reef is elevated 6–8~m above the surrounding sediment plain, and at Site 5, where vertical rock cliffs of 2–3~m are present. In both cases, a single global distance threshold is either overly restrictive or overly permissive, because spatial proximity alone is a poor proxy for visual overlap.

Our approach of estimating 3D image footprints from range maps and linking views based on footprint overlap offers several advantages for ground-truthing visual localization results for near-nadir underwater imagery, because it explicitly accounts for vehicle attitude, altitude, and local terrain structure. The method handles cases where range varies across the field of view, unlike footprint models that use a single range and assume a flat seafloor \citep{eustice_visually_2008, mahon_efficient_2008}. Because the linking criterion in \Cref{eq:camera_footprint_linking} operates directly on footprints and their \gls{IOU}, it compensates for variations in relative geometry between views. The main drawback is the added complexity of requiring calibrated cameras, accurate 3D poses, and reliable range estimates.

For this dataset, the footprint-based ground-truthing approach yields a conservative ground truth with relatively few links per query. Nonetheless, by combining a principled footprint-based linking criterion with a deliberately conservative \gls{IOU} threshold, we are confident that linked camera views exhibit meaningful visual overlap despite residual registration errors and varying camera–seafloor geometry, making the resulting ground truth suitable for benchmarking long-term visual localization algorithms. Future work could refine the footprint estimates by sampling more image points in \Cref{eq:image_footprint_local} to obtain more detailed 3D footprint polygons, or by using more advanced range map estimation methods \citep{ganj_hybriddepth_2025, zhang_spade_2025}.


\subsection{Evaluating Long-Term Visual Place Recognition Performance}
\label{sec:discussion_benchmarking_long_term_visual_place_recognition}

The results in \Cref{sec:results_evaluating_vpr_performance_across_sites} show that the Recall@K of state-of-the-art \gls{VPR} models is significantly lower on our dataset than on comparable terrestrial benchmarks \citep{ali-bey_mixvpr_2023, izquierdo_optimal_2024, berton_megaloc_2025}, highlighting the challenges of long-term \gls{VPR} in benthic environments. Performance varies substantially across sites, with higher recall for Sites 1, 3, and 4, which are dominated by more textured seafloor types. MegaLoc is the strongest model on our dataset, with AnyLoc second, and the \gls{VIT}-based models (AnyLoc, CliqueMining, MegaLoc, SALAD) consistently outperform the \gls{CNN}-based models (CosPlace, EigenPlaces, MixVPR, NetVLAD).

For comparison with other underwater benchmarks, we consider the Eiffel Tower dataset, which is currently the only other curated dataset for long-term visual localization in benthic environments \citep{boittiaux_eiffel_2023}. \gls{VPR} performance on Eiffel Tower has been reported for AnyLoc, CosPlace, MegaLoc, MixVPR, and NetVLAD \citep{gorry_image-based_2025}. Across most of these models, Recall@1 and Recall@10 are lower on our dataset than on Eiffel Tower. For example, MegaLoc is reported to achieve 33.7\% Recall@1 and 65.1\% Recall@10 on Eiffel Tower, compared to 21.9\% and 46.3\% at best on our sites \citep{gorry_image-based_2025}. In contrast to Eiffel Tower, our dataset covers multiple sites with varied seafloor types and photic‑zone benthic habitats, which exhibit stronger temporal dynamics and biological variability \citep{wong_systematic_2023, perkins_temporal_2025, glover_temporal_2010}. Taken together, the consistently lower Recall@K scores and the pronounced cross‑site variability indicate that our benchmark exposes a more diverse and generally harder set of long‑term \gls{VPR} and visual localization challenges.

However, aggregate metrics obscure how performance depends on local seafloor features within each site. Visualizing the spatial distribution of successful and unsuccessful place recognitions for individual visit pairs (\Cref{fig:localization_map_site1_2010_2013,fig:localization_map_site2_2011_2013}) provides additional insight into how seafloor characteristics and motion patterns shape \gls{VPR} behavior. Across sites, successful place recognitions are not uniformly distributed but occur in spatial clusters that coincide with seafloor patches containing visually distinctive and persistent features across visits (for example, dense coral cover, rock–sand interfaces, or boulders with unique geometry) and with trajectory segments with sufficient visual overlap between query and database images. In contrast, visually homogeneous areas with limited persistent features, such as extensive soft‑sediment plains, contain fewer and more sporadically distributed clusters of successful recognitions, although sporadic colonies of sessile organisms (for example, seagrass meadows) can still provide enough visual cues to recognize places.

Site‑level differences in Recall@K (\Cref{tab:mean_place_recognition_recall}) therefore mainly reflect the availability and spatial distribution of such “good” regions for \gls{VPR}. At Site 1, clusters of recognized queries occur frequently and are relatively evenly distributed over the dense coral reef, leading to higher mean recall, with similar behavior at Sites 3 and 4 over textured seafloors. At Site 2, the sharp contrast between coral reef, transition zones with sparse corals, and soft‑sediment bottom produces an uneven pattern, where recognized queries concentrate over coral colonies and other localized features and are rare over homogeneous sediment, lowering overall Recall@K despite some locally good performance. This suggests that, in benthic environments, reliable long-term single-image \gls{VPR} is inherently restricted to localized parts of each habitat where seafloor appearance is distinctive. Sufficient visual overlap between query and database images then becomes a second necessary condition for robust performance.

In \Cref{sec:results_vpr_performance_across_revisits}, we reported how Recall@10 changes with revisit interval. Overall, Recall@10 decreases with increasing revisit interval, consistent with growing scene change between visits, with Site 1 providing the clearest evidence for temporal trends. For both Site 1 and Site 2, \Cref{fig:recall@10_vs_revisit_interval_scatter_plot} shows a marked drop in Recall@10 when the revisit interval increases from 1 to 2 years, followed by a more gradual decline that appears to approach a plateau at longer intervals. This supports the view that most of the visually relevant scene change for \gls{VPR} occurs within the first few years after acquisition, after which additional change has a smaller marginal effect on performance. However, only Sites 1 and 2 have four visits and Site 2 has irregular revisit intervals, and visit-to-visit differences in coverage and altitude (\Cref{tab:site_visit_pair_geometric_statistics}) confound the interpretation, so the observed degradation should be viewed qualitatively rather than as a precise estimate. Despite these limitations, our results indicate that single-image \gls{VPR} is most reliable for relatively short revisit intervals on the order of 1–2 years, and that longer intervals rapidly decrease performance, which has direct consequences for designing long-term monitoring campaigns and revisit schedules.

While the mean Recall@K reported in \Cref{sec:results_evaluating_vpr_performance_across_sites} is useful to compare model performance, practical \gls{VPR} deployment on underwater robots must account for computational and memory constraints. \gls{VIT}-based models generally require more computation than \gls{CNN}-based models \citep{cordonnier_relationship_2020}. Within this group, AnyLoc is particularly expensive, producing 49,152-dimensional descriptors, whereas CliqueMining, MegaLoc, and SALAD use 8,448-dimensional descriptors, resulting in a substantially smaller database footprint and faster retrieval. Considering both retrieval performance and resource usage, MegaLoc is a reasonable default choice for systems that can support \gls{VIT}-based models, whereas EigenPlaces offers the best trade-off among the \gls{CNN}-based methods for platforms that cannot accommodate transformer-based architectures.

Our study has several limitations that qualify these findings. Constructing long‑term benthic datasets with precisely registered imagery is labor‑intensive, which constrains both the number of sites and the diversity of motion patterns and habitat types we can cover. The temporal analysis is dominated by two sites and includes irregular revisit intervals, so the observed degradation over time should be interpreted qualitatively rather than as a precise estimate of performance loss. In addition, we only evaluate single‑image, descriptor‑based \gls{VPR} pipelines without explicit map building or uncertainty estimation, whereas deployed systems will typically combine retrieval with additional localization and fusion modules.

To improve long-term visual localization performance in benthic environments, future work could exploit our observations of clustered successful recognitions by exploring map representations that decompose habitats into sub-maps or place representations formed from spatially and visually coherent image clusters \citep{li_underwater_2015, steinberg_towards_2010, steinberg_hierarchical_2015, liang_online_2025}. Such map representations could reduce the effective search space during retrieval, expose which visual features support successful localization, and incorporate priors on the expected persistence of appearance at each place \citep{johns_images_2011}. Another promising direction is to use information from multiple query images along trajectory segments \citep{stenborg_using_2020, toft_long-term_2022}, increasing the spatial context used for localization and reducing reliance on high overlap between individual query–database image pairs. Accurate short-term odometry from dead-reckoning navigation sensors \citep{fogh_sorensen_localization_2025} can provide relative pose estimates between query images, which can then be used to geometrically reject or verify \gls{VPR} candidates \citep{claxton_improving_2024}.


\subsection{Evaluating Differences In Ground Truth Strategies}
\label{sec:discussion_evaluating_differences_in_ground_truth_strategies}

The comparison between footprint-based and location-based ground-truthing strategies in \Cref{sec:results_evaluating_differences_in_ground_truthing_strategies} shows that \gls{VPR} performance metrics are sensitive to how “relevance” is defined. The location-based ground truth increases the average number of links per query (\gls{ALQ}) relative to the footprint-based definition, and because our \gls{VPR} recall R@K is non-decreasing as additional ground-truth links are added, switching to the location-based ground truth systematically raises R@10. This can give an overly optimistic impression of performance by rewarding any retrieval within a generous distance radius, and similar effects have been reported for dense underwater datasets where overly permissive thresholds allow random baselines to approach \gls{SOTA} Recall@K at moderate rank cutoffs ($K \gtrsim 10$) \citep{gorry_image-based_2025}. In contrast, the information-retrieval definition IR@K normalizes by the total number of relevant links and therefore decreases when the ground truth is expanded but many of the additional links are not retrieved, as observed in our experiments. The larger drop in IR@10 for the stronger \gls{VPR} models in \Cref{fig:results_ground_truth_comparison_scatter} indicates that these models expose inconsistencies between the ground-truth definitions more clearly.

As discussed in \Cref{sec:discussion_linking_camera_views_with_overlapping_footprints}, terrain relief and vehicle altitude variations at Sites 2, 4, and 5 cause substantial differences between footprint-based and location-based ground truths because spatial proximity is a poor proxy for visual overlap in these settings. At sites with little relief and relatively constant altitude, such as Site~1, the effective image footprint width is approximately constant, so the location-based ground truth can accurately approximate the footprint-based ground truth, and the two definitions yield similar \gls{ALQ} values with only small changes in R@10 and IR@10. In contrast, at sites with strong vertical transitions or highly rugged terrain, a relatively loose distance threshold can include many view pairs with limited or no visual overlap, inflating the number of links and amplifying the discrepancy between R@10 and IR@10. Under these conditions, a single global distance threshold cannot capture geometric variations between views and will be overly restrictive in some cases and overly permissive in others.

These findings suggest that overlap-aware ground truths are preferable to purely distance-based criteria for densely mapped near-nadir imagery, since global distance thresholds cannot account for terrain relief, ruggedness, or variations in vehicle altitude. They also show that the \gls{VPR} definition of recall, R@K, is inherently favorable to permissive ground-truth definitions, and that additional metrics such as the information-retrieval definition of recall or mean average precision \citep{garg_where_2021} should be reported to provide more nuanced insight that can be aligned with task- and domain-specific requirements. A limitation of our study is that we only evaluate a single location-based threshold per visit pair. Future work could investigate how performance metrics change as the location threshold is varied over a range, including values that substantially exceed the typical distances between footprint-based view links.


\section{Conclusion}
\label{sec:conclusion}

In this work, we addressed the lack of curated datasets for long-term visual localization in benthic environments and the lack of ground-truthing methods suited for near‑nadir underwater imagery. We presented a curated \gls{AUV} imagery dataset from five benthic reference sites, including raw and color-corrected stereo imagery, camera calibrations, and camera poses registered to sub-decimeter accuracy. To our knowledge, this is the first curated underwater dataset for long-term visual localization that covers photic-zone habitats across multiple sites with diverse seafloor characteristics. Building on this dataset, we proposed a footprint-based ground-truthing method for near-nadir imagery that estimates 3D image footprints and links camera views whose footprints overlap, accounting for terrain relief and vehicle altitude variations and yielding conservative but visually meaningful ground-truth links.

Using this footprint-based ground truth, we benchmarked eight state-of-the-art visual place recognition (\gls{VPR}) methods and found that recall on our dataset is substantially lower than on terrestrial and existing underwater benchmarks, indicating that the sites in our dataset present a diverse and harder set of challenges. We found that performance varied strongly across and within sites, with successful recognitions clustering in regions with distinctive, persistent seafloor features. Comparing our footprint-based ground truth to a traditional location-based ground truth shows that spatial proximity alone can systematically overestimate Recall@K, especially at sites with rugged terrain or large altitude variations, whereas the footprint-based ground truth is more aligned with actual \gls{VPR} retrievals. These findings highlight that accurate ground-truthing of visual localization for near-nadir underwater imagery requires precise geometric information and overlap-aware definitions of “place”, and that reliable long-term visual localization in benthic environments will likely require more sophisticated approaches than single-image \gls{VPR} alone.


\newpage


{
  \setglossarystyle{tree-cap-desc}%
  \renewcommand*{\glsnamefont}[1]{#1}
  \printglossary[type=\acronymtype, title=Acronyms]
}


\section*{Acknowledgment}

The authors would like to acknowledge the Australian Centre of Field Robotics (ACFR) for providing access to the raw data collected by AUV Sirius. The AUV facility at the ACFR that collected the data used in this paper was supported by Australia’s Integrated Marine Observing System (IMOS). IMOS is enabled by the National Collaborative Research Infrastructure Strategy (NCRIS). It is operated by a consortium of institutions as an unincorporated joint venture, with the University of Tasmania as Lead Agent.

The authors acknowledge the use of Perplexity AI (version 3; developed by Perplexity AI, Inc., accessed at \url{https://www.perplexity.ai}) to improve the clarity and readability of the manuscript. The tool was used to refine language and phrasing, but the authors take full responsibility for the content and interpretation of the work.

Base map data © Commonwealth of Australia (Australian Bureau of Statistics) 2021, Australian Statistical Geography Standard (ASGS) Edition 3 digital boundaries, used under Creative Commons Attribution 4.0 International licence.


\bibliographystyle{unsrt}
\bibliography{references}


\newpage
\appendix

\section*{Supplementary Material}
\setcounter{section}{0}  

\section{Supplementary Tables}


\begin{table}[!ht]
    \centering
    \footnotesize
    \begin{tabular}{ccccc}
        \toprule
        Registration Stage & Estimation Method & Parameters & Configuration 1 & Configuration 2 \\
        \midrule
            \multirow{2}{*}{Stage 1}
            & \multirow{2}{*}{FPFH-RANSAC}
            & Voxel Size & 0.08 m & 0.20 m \\
            & & Distance Threshold & 0.06 m & 0.15 m \\
        \midrule
            \multirow{2}{*}{Stage 2}
            & \multirow{2}{*}{Colored ICP}
            & Voxel Size & 0.10 m & 0.20 m \\
            & & Distance Threshold & 0.15 m & 0.20 m \\
        \midrule
            \multirow{2}{*}{Stage 3}
            & \multirow{2}{*}{Colored ICP}
            & Voxel Size & 0.05 m & 0.05 m \\
            & & Distance Threshold & 0.05 m & 0.05 m \\
        \midrule
            \multirow{2}{*}{Stage 4}
            & \multirow{2}{*}{Colored ICP}
            & Voxel Size & 0.02 m & 0.02 m \\
            & & Distance Threshold & 0.02 m & 0.02 m \\
        \bottomrule \\
    \end{tabular}
    \caption{
        The two configurations of the geometric registration pipeline that is used to register the visits for each site in our dataset. Configuration 1 is used to register the visits for Site 1, while Configuration 2 is used to register the visits to Site 2, 3, 4, and 5. Overall, Configuration 1 performs registration at finer spatial resolutions than Configuration 2.
    }
    \label{tab:registration_pipeline_configuration}
\end{table}


\newcommand{\methodLabelRotation}{0}
\newcommand{\ruleSpacing}{2pt}
\setlength{\tabcolsep}{2.9pt} 


\begin{table}[!ht]
    \centering
    \footnotesize
    \begin{tabular}{cccc cccc cccc cccc cc}
    \toprule
    \multirow[c]{2}{*}{\centering Site}
    & \multirow[c]{2}{*}{\centering Visit Years}
    & \multicolumn{2}{c}{\rotatebox{\methodLabelRotation}{AnyLoc}}
    & \multicolumn{2}{c}{\rotatebox{\methodLabelRotation}{CliqueMining}}
    & \multicolumn{2}{c}{\rotatebox{\methodLabelRotation}{CosPlace}}
    & \multicolumn{2}{c}{\rotatebox{\methodLabelRotation}{EigenPlaces}}
    & \multicolumn{2}{c}{\rotatebox{\methodLabelRotation}{MegaLoc}}
    & \multicolumn{2}{c}{\rotatebox{\methodLabelRotation}{MixVPR}}
    & \multicolumn{2}{c}{\rotatebox{\methodLabelRotation}{NetVLAD}}
    & \multicolumn{2}{c}{\rotatebox{\methodLabelRotation}{SALAD}}
    \\
    \cmidrule(l{\ruleSpacing}r{\ruleSpacing}){3-4}
    \cmidrule(l{\ruleSpacing}r{\ruleSpacing}){5-6}
    \cmidrule(l{\ruleSpacing}r{\ruleSpacing}){7-8}
    \cmidrule(l{\ruleSpacing}r{\ruleSpacing}){9-10}
    \cmidrule(l{\ruleSpacing}r{\ruleSpacing}){11-12}
    \cmidrule(l{\ruleSpacing}r{\ruleSpacing}){13-14}
    \cmidrule(l{\ruleSpacing}r{\ruleSpacing}){15-16}
    \cmidrule(l{\ruleSpacing}r{\ruleSpacing}){17-18}
    & 
    & R@1 & R@10
    & R@1 & R@10
    & R@1 & R@10
    & R@1 & R@10
    & R@1 & R@10
    & R@1 & R@10
    & R@1 & R@10
    & R@1 & R@10
    \\
    \midrule
    \multirow[c]{6}{*}{\centering Site 1}
    & 2010-2011 
        & \bestRecall{42.3} & \bestRecall{73.0} & 30.3 & 56.2 
        & 20.1 & 49.1 & 21.0 & 49.3 
        & \nextBestRecall{31.3} & \nextBestRecall{59.5} & 12.6 & 35.7 
        & 5.3 & 29.1 & 30.4 & 54.8 \\
    & 2010-2012 
        & \bestRecall{20.9} & \bestRecall{50.9} & 14.4 & 40.0 
        & 14.0 & 36.2 & 13.2 & 37.4 
        & \nextBestRecall{19.2} & \nextBestRecall{47.2} & 6.2 & 22.1 
        & 2.9 & 18.3 & 15.2 & 41.2 \\
    & 2010-2013 
        & \bestRecall{15.2} & \bestRecall{42.3} & 10.8 & 34.1 
        & 10.3 & 33.0 & 12.4 & 35.6 
        & 13.1 & 36.3 & 7.4 & 28.8 & 4.5 & 20.8 
        & \nextBestRecall{13.7} & \nextBestRecall{37.2} \\
    & 2011-2012 
        & \bestRecall{24.2} & \bestRecall{47.6} & 16.7 & 39.0 
        & 11.6 & 32.9 & 12.6 & 33.4 
        & \nextBestRecall{23.0} & \nextBestRecall{46.1} & 6.6 & 21.3 
        & 3.6 & 16.8 & 19.4 & 40.8 \\
    & 2011-2013 
        & \bestRecall{14.2} & \bestRecall{35.7} & 8.3 & 27.6 
        & 10.2 & 27.7 & 10.9 & 29.2 
        & \nextBestRecall{13.8} & \nextBestRecall{33.9} & 4.7 & 19.7 
        & 4.8 & 18.5 & 10.0 & 28.8 \\
    & 2012-2013 
        & \bestRecall{32.5} & \bestRecall{57.7} & 21.9 & 42.1 
        & 16.7 & 43.5 & 18.2 & 43.2 
        & \nextBestRecall{31.1} & \nextBestRecall{54.7} & 9.1 & 26.3 
        & 4.0 & 20.3 & 22.4 & 42.1 \\
    \midrule
    \multirow[c]{6}{*}{\centering Site 2}
    & 2011-2012 
        & \nextBestRecall{9.5} & \bestRecall{32.7} & 5.9 & 28.0 
        & 4.8 & 23.8 & 6.9 & 30.6 
        & \bestRecall{11.1} & \nextBestRecall{32.4} & 2.2 & 16.7 
        & 2.9 & 16.9 & 6.9 & 29.9 \\
    & 2011-2013 
        & \nextBestRecall{7.2} & \bestRecall{28.7} & 5.8 & 23.1 
        & 3.4 & 16.4 & 3.5 & 20.1 
        & \bestRecall{8.2} & \nextBestRecall{27.6} & 2.7 & 15.5 
        & 2.3 & 14.7 & 5.5 & 23.9 \\
    & 2011-2017 
        & \nextBestRecall{5.5} & \bestRecall{26.1} & 3.6 & 21.0 
        & 3.3 & 17.5 & 3.7 & 18.1 
        & \bestRecall{5.8} & \nextBestRecall{22.7} & 1.9 & 17.4 
        & 3.3 & 14.4 & 3.8 & 20.9 \\
    & 2012-2013 
        & 7.7 & 27.8 & 8.0 & 26.1 
        & 4.6 & 18.9 & 5.6 & 22.7 
        & \bestRecall{10.3} & \bestRecall{28.9} & 3.3 & 18.4 
        & 2.8 & 16.7 & \nextBestRecall{8.0} & \nextBestRecall{27.8} \\
    & 2012-2017 
        & \nextBestRecall{6.8} & \bestRecall{24.8} & 4.6 & 23.1 
        & 3.7 & 19.1 & 4.6 & 22.8 
        & \bestRecall{7.0} & 23.9 & 3.1 & 19.1 
        & 2.2 & 13.9 & 6.1 & \nextBestRecall{24.0} \\
    & 2013-2017 
        & \nextBestRecall{7.7} & \nextBestRecall{29.1} & 5.9 & 26.0 
        & 6.3 & 24.6 & 5.0 & 22.9 
        & \bestRecall{9.8} & \bestRecall{30.8} & 2.4 & 16.4 
        & 3.3 & 20.0 & 6.6 & 24.4 \\
    \midrule
    \multirow[c]{3}{*}{\centering Site 3}
    & 2010-2012 
        & \nextBestRecall{16.8} & \nextBestRecall{42.5} & 13.2 & 35.8 
        & 7.2 & 26.5 & 7.9 & 27.5 
        & \bestRecall{23.7} & \bestRecall{49.3} & 3.3 & 13.4 
        & 2.5 & 12.3 & 13.5 & 36.8 \\
    & 2010-2014 
        & 6.7 & 25.6 & 8.4 & 26.0 
        & 3.6 & 19.0 & 3.7 & 20.1 
        & \bestRecall{13.6} & \bestRecall{37.2} & 1.4 & 10.4 
        & 1.7 & 9.3 & \nextBestRecall{9.1} & \nextBestRecall{31.9} 
        \\
    & 2012-2014 
        & 6.5 & 24.4 & 11.1 & 34.9 
        & 5.7 & 23.2 & 6.2 & 23.7 
        & \bestRecall{22.1} & \bestRecall{49.4} & 1.4 & 12.2 
        & 2.5 & 13.4 & \nextBestRecall{13.8} & \nextBestRecall{35.2} 
        \\
    \midrule
    \multirow[c]{3}{*}{\centering Site 4}
    & 2009-2011 
        & \nextBestRecall{24.8} & \bestRecall{57.6} & 14.2 & 33.9 
        & 5.3 & 19.9 & 6.2 & 22.2 
        & \bestRecall{27.3} & \nextBestRecall{56.2} & 5.9 & 25.4 
        & 4.1 & 21.2 & 21.9 & 48.1 
        \\
    & 2009-2013 
        & 21.0 & \nextBestRecall{48.1} & 17.0 & 38.8 
        & 6.0 & 21.1 & 6.6 & 25.4 
        & \bestRecall{23.3} & 47.9 & 3.4 & 15.7 
        & 3.1 & 14.9 & \nextBestRecall{22.8} & \bestRecall{49.9} 
        \\
    & 2011-2013 
        & \bestRecall{18.4} & \bestRecall{44.2} & 9.4 & 24.6 
        & 3.5 & 16.2 & 5.2 & 22.8 
        & \nextBestRecall{12.1} & \nextBestRecall{31.9} & 2.5 & 16.8 
        & 2.3 & 11.7 & 9.5 & 27.3 
        \\
    \midrule
    \multirow[c]{3}{*}{\centering Site 5}
    & 2010-2012 
        & \nextBestRecall{7.8} & \nextBestRecall{27.7} & 4.7 & 19.9 
        & 6.0 & 23.9 & 4.9 & 20.5 
        & \bestRecall{9.4} & \bestRecall{28.6} & 3.6 & 17.2 
        & 2.2 & 13.0 & 7.3 & 24.4 
        \\
    & 2010-2013 
        & 8.5 & 28.3 & 6.5 & 24.5 
        & 4.7 & 18.5 & 5.4 & 19.7 
        & \bestRecall{10.5} & \bestRecall{31.9} & 3.6 & 17.0 
        & 2.3 & 14.4 & \nextBestRecall{10.1} & \nextBestRecall{30.2} 
        \\
    & 2012-2013 
        & \nextBestRecall{13.9} & \nextBestRecall{39.6} & 10.4 & 32.4 
        & 6.2 & 24.6 & 7.6 & 27.9 
        & \bestRecall{17.9} & \bestRecall{42.2} & 4.5 & 20.5 
        & 3.4 & 19.0 & 13.2 & 34.2 
        \\
    \bottomrule \\
    \end{tabular}
    \caption{
        Recall@1 and Recall@10 for all VPR models and visit pairs in the dataset.
        For each rank cutoff $K$ and visit pair, we use \bestRecall{green} and \nextBestRecall{red} to highlight the best and second best model, respectively.
    }
    \label{tab:recall_all_models_all_visit_pairs}
\end{table}


\newpage

\section{Supplementary Figures}


\begin{figure}[!ht]
    \centering
    \includegraphics[width=\textwidth]{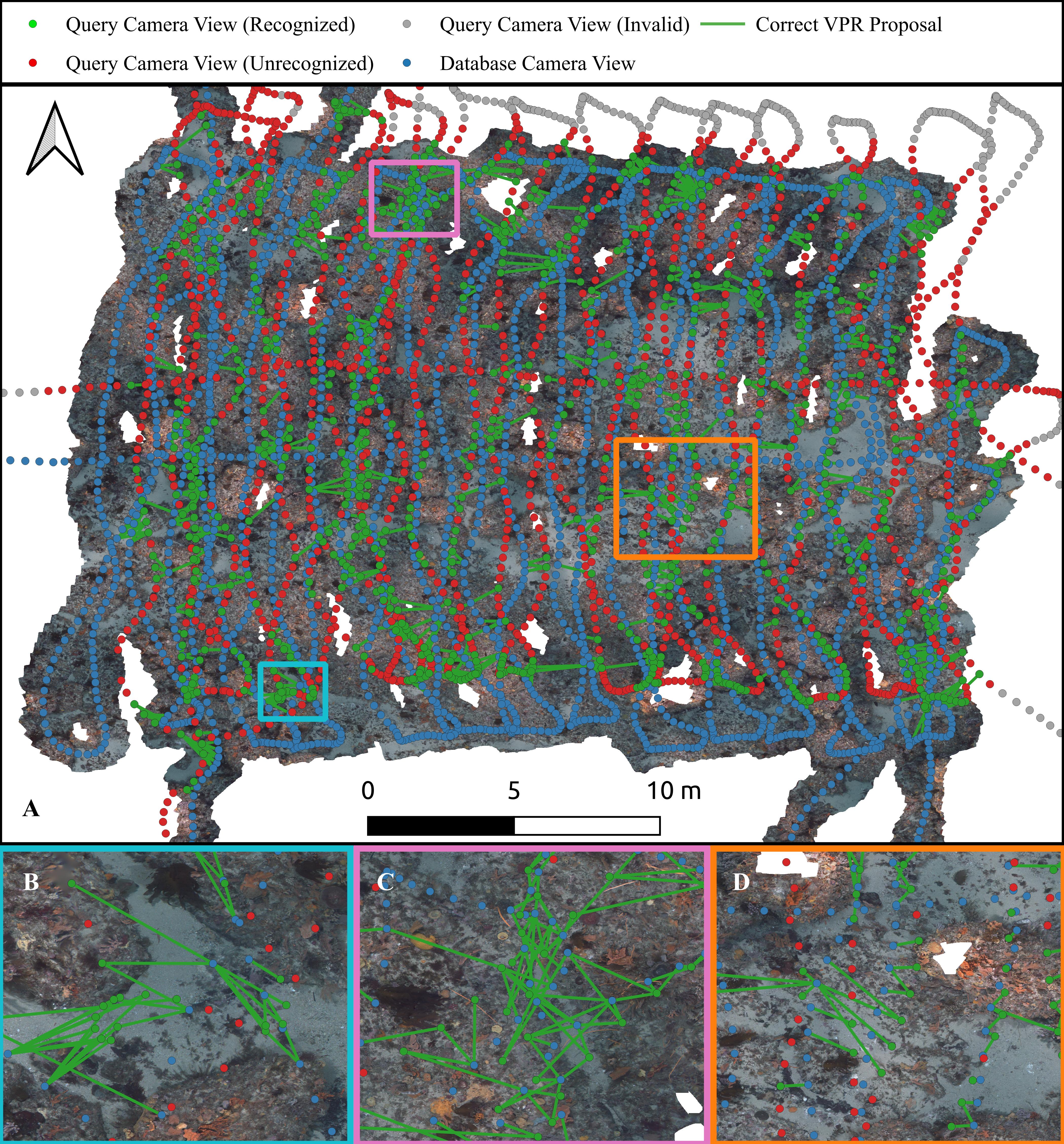}
    \caption{
        Visual place recognition results at Site 3 rendered on top of an orthomosaic derived from the 2010 images, using MegaLoc with $K=5$ to retrieve 2010 database images for 2012 query images.
    }
    \label{fig:localization_map_site3_2010_2012}
\end{figure}


\newpage

\begin{figure}[!ht]
    \centering
    \includegraphics[width=\textwidth]{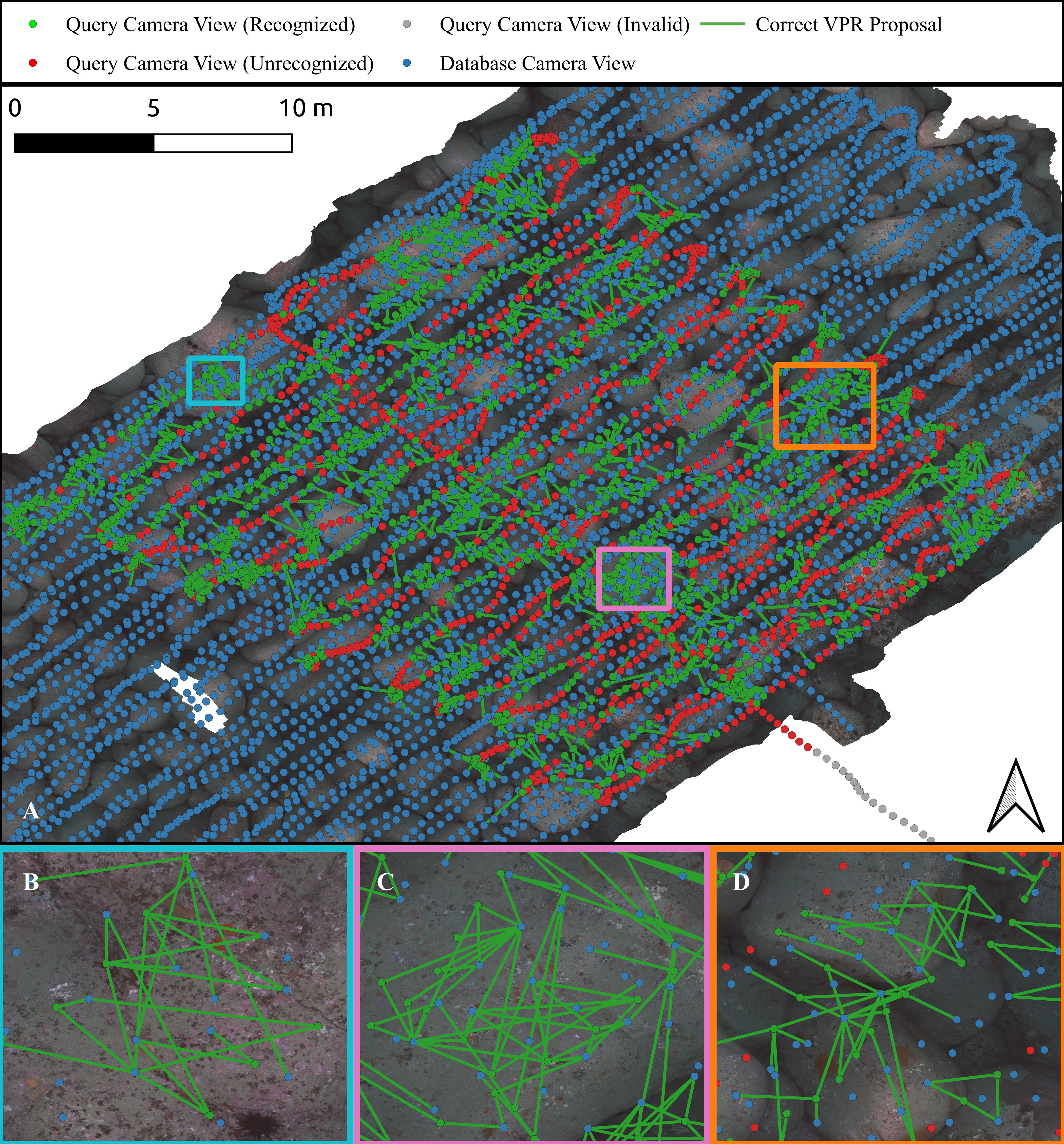}
    \caption{
        Visual place recognition results at Site 4 rendered on top of an orthomosaic derived from the 2009 images, using MegaLoc with $K=5$ to retrieve 2009 database images for 2011 query images.
    }
    \label{fig:localization_map_site4_2009_2011}
\end{figure}


\newpage

\begin{figure}[!ht]
    \centering
    \includegraphics[width=\textwidth]{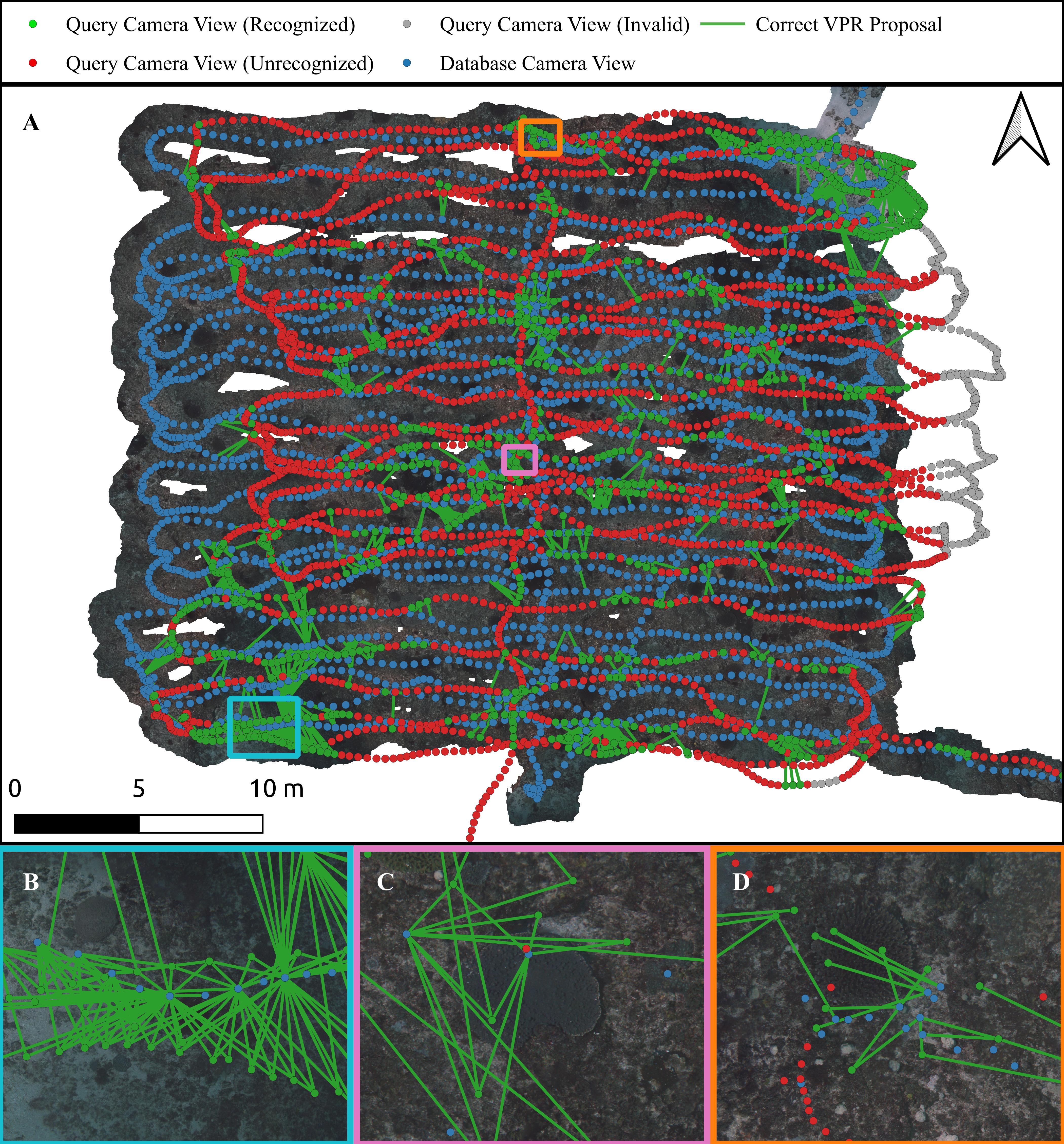}
    \caption{
        Visual place recognition results at Site 5 rendered on top of an orthomosaic derived from the 2010 images, using MegaLoc with $K=5$ to retrieve 2010 database images for 2013 query images.
    }
    \label{fig:localization_map_site5_2010_2013}
\end{figure}


\newpage

\begin{figure}[!ht]
    \centering
    \includegraphics[width=0.75\textwidth]{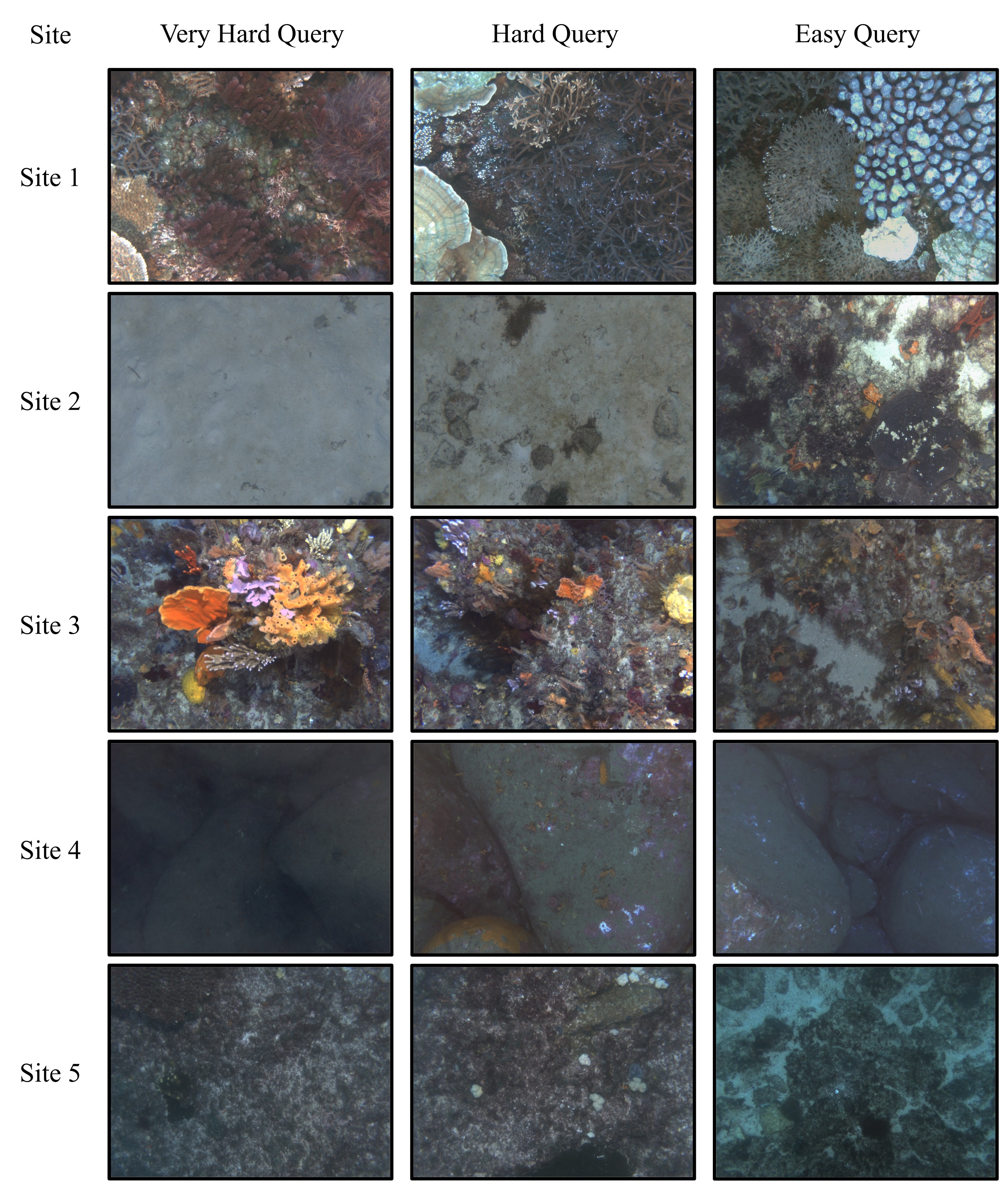}
    \caption{
        Examples of very hard, hard, and easy query images from each benthic reference site. The very hard queries are not recognized by any VPR model, hard queries are recognized by the best models, i.e. MegaLoc and AnyLoc, and easy queries are recognized by all models.
    }
    \label{fig:query_examples}
\end{figure}


\newpage

\begin{figure}[!ht]
    \centering
    \includegraphics[width=\textwidth]{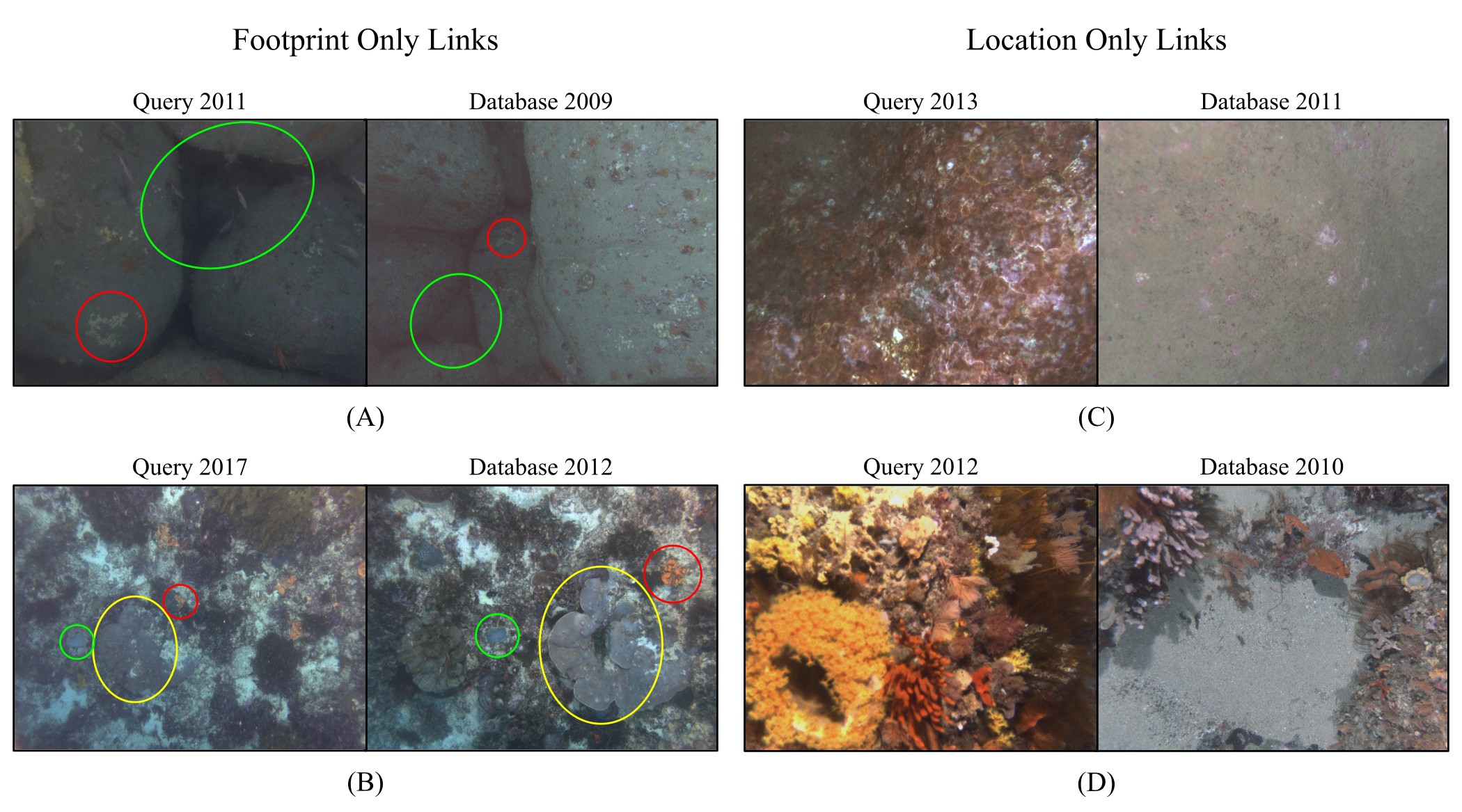}
    \caption{
        Examples illustrating disagreement between footprint-based and location-based linking of query–database camera views. Location-based linking used a threshold corresponding to the 95th percentile of spatial distances between footprint-linked camera views. Panels~(A) and~(B) show image pairs that are linked by overlapping footprints but not by location proximity. These cases involve large altitude differences between the query and database camera views, preventing location-based linking. Manually added markers highlight visual features shared between the paired images. Panels~(C) and~(D) show image pairs linked by location proximity but not by overlapping footprints. Here, abrupt changes in seafloor relief cause the cameras to operate at low altitudes, limiting the field of view. Although these image pairs are spatially close, they lack shared visual features.
    }
    \label{fig:camera_view_link_disagreement_examples}
\end{figure}

\end{document}